\documentclass[11pt]{article}

\usepackage[preprint]{acl}

\usepackage{enumitem}
\usepackage{times}
\usepackage{latexsym}

\usepackage[T1]{fontenc}

\usepackage[utf8]{inputenc}

\usepackage{microtype}

\usepackage{inconsolata}

\usepackage{amsmath}
\usepackage{amssymb}
\usepackage{mathtools}
\usepackage{amsthm}
\usepackage{algorithm}
\usepackage{booktabs}
\usepackage{makecell}
\usepackage{multirow}
\usepackage[table]{xcolor}
\usepackage{threeparttable}
\usepackage{graphicx}
\usepackage{subcaption}
\usepackage{fvextra}
\usepackage{algpseudocode}

\usepackage[most]{tcolorbox}
\tcbuselibrary{breakable,skins,listings}
\usepackage{listings}

\definecolor{promptbg}{RGB}{248,250,252}
\definecolor{promptframe}{RGB}{99,121,145}
\definecolor{prompttitlebg}{RGB}{38,70,105}
\definecolor{prompttitlefg}{RGB}{255,255,255}
\definecolor{prompttext}{RGB}{35,45,55}

\lstdefinestyle{promptstyle}{
    basicstyle=\ttfamily\scriptsize\color{prompttext},
    columns=fullflexible,
    breaklines=true,
    breakatwhitespace=false,
    keepspaces=true,
    showstringspaces=false,
    tabsize=2,
    upquote=true
}

\newtcblisting{promptbox}[1]{
    enhanced,
    breakable,
    listing only,
    listing engine=listings,
    listing options={style=promptstyle},
    colback=promptbg,
    colframe=promptframe,
    colbacktitle=prompttitlebg,
    coltitle=prompttitlefg,
    title={\parbox[t]{\dimexpr\linewidth-6mm\relax}{\raggedright #1}},
    fonttitle=\sffamily\bfseries\small,
    boxrule=0.6pt,
    arc=1.5mm,
    left=1.2mm,
    right=1.2mm,
    top=1.2mm,
    bottom=1.2mm,
    toptitle=1mm,
    bottomtitle=1mm,
    lefttitle=1.5mm,
    righttitle=1.5mm,
    before skip=0.8em,
    after skip=0.8em
}

\definecolor{mylightblue}{rgb}{0.8,0.9,1.0}
\definecolor{mylighterblue}{rgb}{0.9,0.95,1.0}

\title{Co-Evolving LLM Evaluators and Policies via DynamicRubric}

\author{
  \textbf{Beining Wang}$^{1,2,\dagger}$ \quad
  \textbf{Weihang Su}$^{1}$ \quad
  \textbf{Hongtao Tian}$^{2}$ \quad
  \textbf{Hao Kong}$^{2}$ \\
  \textbf{Tao Yang}$^{2}$ \quad
  \textbf{Ting Yao}$^{2}$ \quad
  \textbf{Qingyi Pan}$^{1}$ \quad
  \textbf{Yueyue Wu}$^{1}$ \\
  \textbf{Qingyao Ai}$^{1,\ddagger}$ \quad
  \textbf{Min Zhang}$^{1}$ \quad
  \textbf{Yiqun Liu}$^{1}$ \\[0.4cm] 
  $^{1}$Department of Computer Science and Technology, Tsinghua University \\
  $^{2}$WeChat, Tencent
}

\begin{document}
\maketitle
\begingroup
\renewcommand\thefootnote{} 

\footnotetext{\hspace{-1.5em}$^\dagger$Email: \texttt{wbn23@mails.tsinghua.edu.cn}}
\footnotetext{\hspace{-1.5em}$^\ddagger$Corresponding author. Email: \texttt{aiqy@tsinghua.edu.cn}}

\endgroup
\begin{abstract}
Post-training with evaluator feedback on policy-induced samples serves as a major mechanism for improving large language models.
As policies improve, these sampled responses become close in quality.
These close candidates create a bottleneck for policy optimization: collapsed relative evaluator score gaps yield weak or misleading policy supervision.
We theoretically characterize why these gaps matter through a probability allocation view, showing that the directional gain of shifting probability mass from one response to another is exactly the evaluator score gap between them.
This identifies relative score gaps as the policy optimization signals that guide updates.
Motivated by this view, we propose \textbf{DynamicRubric}, a response-set-conditioned evaluator--policy co-evolution framework that generates weighted binary rubric items for each candidate set and aggregates the resulting judgments into response-level scores.
In our experiments with 8B backbones, DynamicRubric improves evaluator performance and provides stronger policy supervision than baselines using a 70B reward model or a 235B static rubric generator.
DynamicRubric-optimized policies also show gains on verifiable reasoning and coding tasks.
\textbf{A DynamicRubric-optimized model is fully deployed in WeChat Search's AI answering scenario, where it serves all online traffic across tens of millions of requests per day and improves key online metrics.}
These results suggest a principle for evaluator-guided post-training: evaluators should evolve with the policies they supervise.
\end{abstract}

\section{Introduction}

\begin{figure*}[t]
\begin{center}
\includegraphics[width=0.9\linewidth]{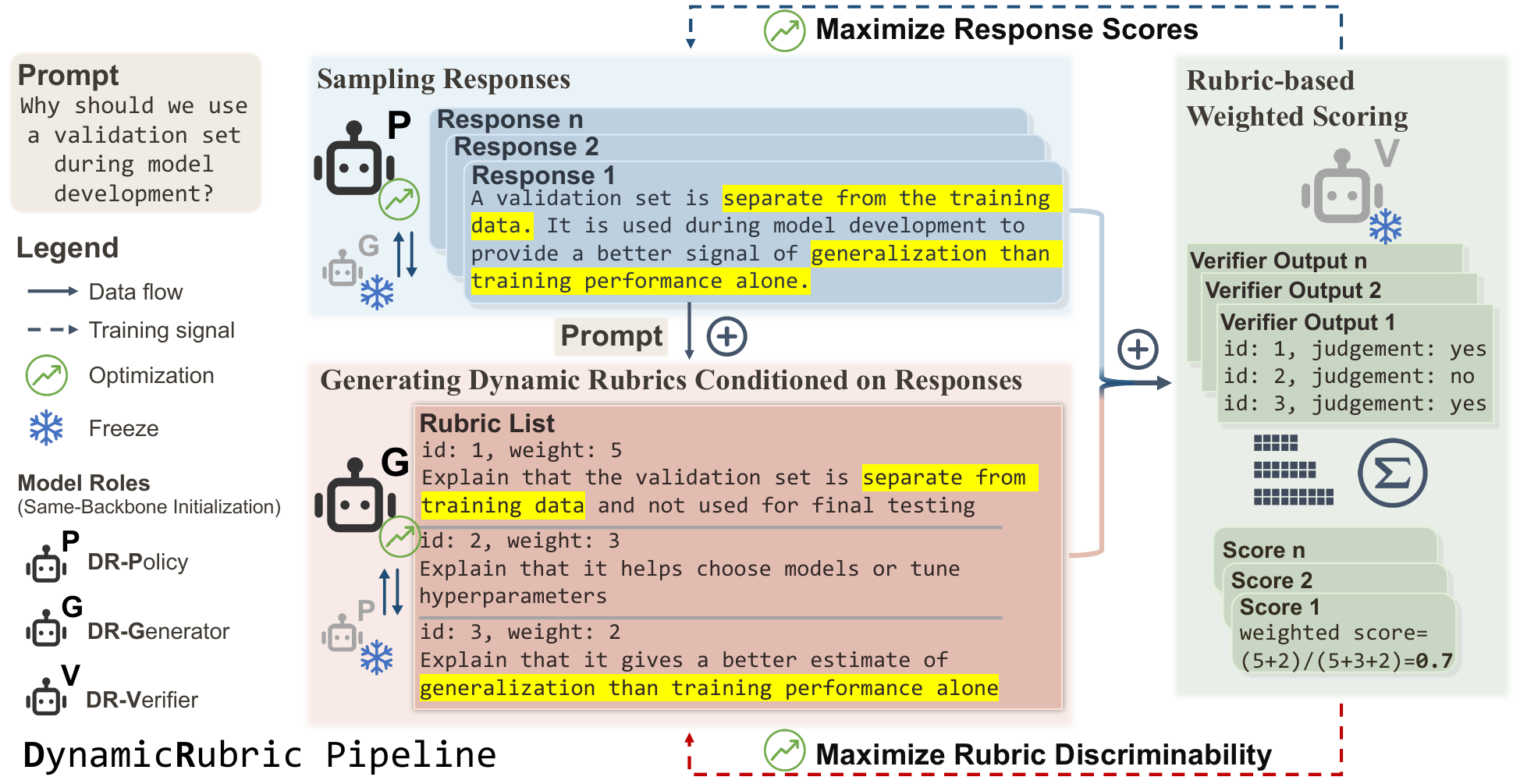}
\end{center}
\caption{
Overview of DynamicRubric.
For each prompt, DR-Policy samples a candidate response set, DR-Generator produces weighted binary rubric items conditioned on the candidate responses, and a frozen DR-Verifier applies the rubrics to compute response-level scores used to co-optimize DR-Policy and DR-Generator.
}
\label{fig:main}
\end{figure*}
Post-training with feedback from evaluators has become a central mechanism for improving large language models (LLMs)~\citep{ouyang2022training,rafailov2023direct}.
In these pipelines, policies are optimized using feedback on responses sampled from the current policy.
As optimization improves a policy, its sampled responses often become close in quality, making the remaining useful differences subtle and instance-specific.
This creates a central bottleneck for evaluator-guided optimization: policy updates depend on relative score gaps among close candidates, which determine how probability mass should be reallocated.
Existing reward models and LLM-as-Judge evaluators are often unreliable for such fine-grained comparisons~\citep{lambert2025rewardbench,tan2025judgebench}.
When these gaps collapse or point in the wrong direction, policy optimization receives weak or misleading feedback.

Rubric-based evaluation offers a natural way to make fine-grained differences explicit by decomposing holistic judgments into criteria~\citep{liu2023geval,ye2023flask,kim2024prometheus,song2024finesure}.
Yet a rubric helps policy optimization only when its criteria separate the responses that the current policy actually produces.
As policies improve, these candidate responses often satisfy broad criteria such as helpfulness and clarity.
Once these broad criteria become saturated, the distinctions that matter for policy optimization become more instance-specific: a missed constraint, a flawed reasoning step, or a choice that fails to match the user's intent.
Rubrics written before observing the candidate responses can easily overlook the distinctions that determine the next policy update.

Taken together, this gap-preservation requirement motivates \emph{dynamic}, response-set-conditioned rubric evaluation.
Given a prompt and responses sampled from the current policy, the evaluator generates binary rubric items aimed at the differences that matter within the set.
A verifier applies these items to every response and aggregates the yes/no judgments into response-level scores.
The shared rubric makes the resulting score gaps directly comparable within the set and traceable to explicit judgments, yielding feedback aligned with the comparisons that drive policy optimization.

We theoretically ground this design through a probability allocation view of policy optimization.
Under this view, conditioned on a prompt and a sampled response set, the directional gain of shifting probability mass from one candidate to another is exactly the evaluator score gap between them.
This identity turns evaluator discriminability into an optimization quantity: relative score gaps are the signals that guide policy updates within the response set.
We further establish that response-set-conditioned evaluators include prompt-only evaluators as a special case, while the response-set input allows the evaluator to adapt to the candidates being compared.
Since policy updates continually change the response distribution, the evaluator must adapt to the response sets it is used to supervise.
Together, these results yield an evaluator--policy co-evolution principle: adapt the evaluator to current-policy response sets, then optimize the policy using the adapted evaluator.

To instantiate this bilevel principle, we propose \textbf{DynamicRubric}, a response-set-conditioned rubric framework.
As illustrated in Figure~\ref{fig:main}, given a prompt and candidate responses sampled from the current policy, \textbf{DR-Generator} produces weighted binary rubric items targeting distinctions within the response set.
\textbf{DR-Verifier} applies each item to each response, producing yes/no judgments that are aggregated by weights into response-level scores.
DynamicRubric alternates between adapting DR-Generator on response sets from the evolving \textbf{DR-Policy} and optimizing DR-Policy using the resulting rubric-based scores as supervision signals.

Experiments show that DynamicRubric improves both evaluator performance and downstream policy optimization.
Using Qwen3-8B as the backbone, the trained DR-Generator outperforms larger evaluator baselines on preference evaluation, including a Qwen3-32B zero-shot dynamic rubric generator.
When used as the supervision signal for policy optimization, DynamicRubric consistently improves open-ended generation over supervision from a 70B reward model and static rubric supervision from a 235B rubric generator.
Policies optimized with DynamicRubric also obtain consistent gains on math, coding and scientific QA benchmarks, suggesting its policy-level benefits extend beyond open-ended generation to verifiable reasoning tasks.
Continuous co-evolution experiments further show that evolved evaluators provide stronger supervision than stale evaluators.
Beyond benchmark evaluation, we deploy a DynamicRubric-optimized model in WeChat Search's AI answering scenario, a large-scale online production setting that serves tens of millions of requests per day.
Online A/B tests against the previous production model show statistically significant improvements in total search volume, user duration, and absolute positive user behaviors.
\textbf{The resulting model is fully deployed in production and currently serves all online traffic, replacing the previous production model.}
Together, these results point to a central requirement for post-training evaluators: they should evolve with the policies to maintain relative discriminability.
DynamicRubric meets this requirement by letting the rubric evaluator evolve with the response sets encountered during optimization, keeping evaluation focused on the comparisons that determine the next policy update.

\section{Related Work}

\subsection{Rubric-Based Evaluation and Feedback}

LLM-based evaluators are widely used to assess open-ended generation and approximate human preferences at scale~\citep{zheng2023judging,liu2023geval}.
Rubric-based evaluation improves interpretability by decomposing response quality into explicit criteria, structured forms, or fine-grained dimensions~\citep{liu2023geval}.
G-Eval uses chain-of-thought and form-filling prompts for natural language generation evaluation, while FLASK decomposes instruction following into fine-grained alignment skills~\citep{liu2023geval,ye2023flask}.
Prometheus trains open-source evaluator models to follow customized grading rubrics and FineSurE designs multi-dimensional fine-grained evaluation for summarization~\citep{kim2024prometheus,song2024finesure}.
Rubrics have also been explored as reward signals for on-policy reinforcement learning beyond domains with deterministic verifiers~\citep{gunjal2025rubrics,zhang2025chasing}.

\subsection{Policy Optimization with Feedback for LLMs}
Post-training with feedback is a standard approach for improving LLM policies.
RLHF trains reward models from human preference comparisons for policy optimization, while RLAIF uses AI-generated feedback to reduce human annotation cost~\citep{ouyang2022training,bai2022constitutional}.
Direct preference optimization further simplifies preference learning by optimizing policies directly from preference pairs without explicitly training a reward model~\citep{rafailov2023direct}.
Recent reasoning-oriented systems extend RL-based post-training with verifiable, outcome-based rewards to improve mathematical, coding, and general reasoning abilities~\citep{shao2024deepseekmath,guo2025deepseek}.
The effectiveness of these methods depends on the reliability of the feedback signal during policy optimization, and imperfect reward models can lead to overoptimization~\citep{gao2023scaling}.
These observations motivate our evaluator-side perspective: post-training should preserve discriminative score gaps on the response distribution being optimized.

\section{Methodology}
\label{sec:methodology}

\subsection{Relative Evaluator Score Gaps as Optimization Signals}
\label{sec:relative-gap}
In LLM post-training, useful evaluators are expected to maintain discriminability among close current-policy responses~\citep{ouyang2022training,zhang2025chasing}.
This section formalizes this requirement by showing that relative evaluator score gaps within policy-induced response sets are exactly the optimization signals.

Formally, let $x\sim\mathcal D$ be a prompt, and let the policy $\pi_\theta(y\mid x)$ denote the conditional generation distribution of the current LLM.
$E(x,y)\in \mathbb R$ denotes the scalar score assigned by an evaluator to a prompt-response pair.
For $x$, we sample $K$ responses from the current policy and denote the sampled response set by $\mathcal C$:
\begin{equation}
\small
\label{eq:sample-response-set}
  y_1,\dots,y_K \stackrel{\mathrm{i.i.d.}}{\sim} \pi_\theta(\cdot\mid x),
  \qquad
  \mathcal C=\{y_1,\dots,y_K\}.
\end{equation}
To characterize the relative score structure that an evaluator should preserve, we introduce an analysis-time pair distribution $P(\cdot\mid x,\mathcal C)$.
This distribution is supported on ordered pairs $(y^+,y^-)$ within $\mathcal C$, where the direction from $y^-$ to $y^+$ is determined by evaluator-independent reference quality.
Define the average \textbf{relative evaluator score gap} on $\mathcal C$ as
\begin{equation}
\small
\label{eq:delta-gap}
\begin{split}
\Delta_{\mathrm{gap}}(E;\theta)
=
\mathbb E_{x\sim\mathcal D}
&\mathbb E_{\mathcal C\sim\pi_\theta(\cdot\mid x)}
\mathbb E_{(y^+,y^-)\sim P(\cdot\mid x,\mathcal C)}\\
&\left[
E(x,y^+)-E(x,y^-)
\right].
\end{split}
\end{equation}
Equation~\ref{eq:delta-gap} measures the average relative score gap that the evaluator exposes on ordered pairs from the current response set.
From a local probability allocation perspective, this gap is also the local optimization signal for moving probability mass from $y^-$ to $y^+$.
To see this, for a fixed $x$ and $\mathcal C$, we define the normalized policy mass on $\mathcal C$ as
\begin{equation}
\small
\label{eq:local-alpha}
\alpha_i=
\frac{\pi_\theta(y_i\mid x)}
{\sum_{j=1}^K \pi_\theta(y_j\mid x)},
\qquad
\sum_{i=1}^K \alpha_i=1.
\end{equation}
In the local coordinates $\boldsymbol\alpha=(\alpha_1,\dots,\alpha_K)$, we consider the local proxy objective
\begin{equation}
\small
\label{eq:local-objective}
\mathcal J^{\mathrm{local}}_{x,\mathcal C}(\boldsymbol\alpha;E)
=
\sum_{i=1}^K \alpha_i E(x,y_i).
\end{equation}

Let $\mathbf d=(d_1,\dots,d_K)$ be any feasible local direction on the probability simplex, satisfying $\sum_{i=1}^K d_i=0$, and define the local path
$\boldsymbol\alpha(\varepsilon)=\boldsymbol\alpha+\varepsilon\mathbf d$.
For sufficiently small $|\varepsilon|$, this path remains inside the simplex.
Since $\mathcal J^{\mathrm{local}}_{x,\mathcal C}$ is linear in $\boldsymbol\alpha$, its directional derivative along this path is
\begin{equation}
\small
\label{eq:local-directional-derivative}
\left.
\frac{\mathrm d}{\mathrm d\varepsilon}
\mathcal J^{\mathrm{local}}_{x,\mathcal C}
(\boldsymbol\alpha+\varepsilon\mathbf d;E)
\right|_{\varepsilon=0}
=
\sum_{i=1}^K d_i E(x,y_i).
\end{equation}
Thus, the local improvement directions on $\mathcal C$ are determined by the relative structure of evaluator scores.
For any pair $(y_i,y_j)$, define the transfer direction
$\mathbf d_{i\leftarrow j}=e_i-e_j$, where $e_i,e_j$ are standard basis vectors.
This direction transfers an infinitesimal amount of probability mass from $y_j$ to $y_i$ while leaving the remaining coordinates unchanged.
The directional derivative along this direction is
\begin{equation}
\small
\label{eq:pair-directional-derivative}
\left.
\frac{\mathrm d}{\mathrm d\varepsilon}
\mathcal J^{\mathrm{local}}_{x,\mathcal C}
(\boldsymbol\alpha+\varepsilon\mathbf d_{i\leftarrow j};E)
\right|_{\varepsilon=0}
=
E(x,y_i)-E(x,y_j).
\end{equation}
Thus, for an ordered pair $(y^+,y^-)$, the relative score gap is exactly the local directional gain of transferring probability mass from $y^-$ to $y^+$.
Since general local updates can be decomposed into such pairwise transfers, relative score gaps characterize the local improvement directions on $\mathcal C$.
When these gaps collapse, the optimization signal exposed by the evaluator shrinks, and policy optimization becomes less effective on the sampled response set.

The above identity holds pointwise for each fixed $x$, $\mathcal C$, and ordered pair $(y^+,y^-)$.
Taking expectation under the same sampling process gives the expected local \textbf{optimization signal}.
Let $\mathbf d_{+\leftarrow -}$ denote the transfer direction from $y^-$ to $y^+$.
Then
\begin{equation}
\small
\label{eq:gap-as-expected-directional-gain}
\begin{split}
\mathbb E_{x\sim\mathcal D}
&\mathbb E_{\mathcal C\sim\pi_\theta(\cdot\mid x)}
\mathbb E_{(y^+,y^-)\sim P(\cdot\mid x,\mathcal C)}\\
&\left[
\left.
\frac{\mathrm d}{\mathrm d\varepsilon}
\mathcal J^{\mathrm{local}}_{x,\mathcal C}
(\boldsymbol\alpha+\varepsilon\mathbf d_{+\leftarrow -};E)
\right|_{\varepsilon=0}
\right]
\\
=&
\mathbb E_{x\sim\mathcal D}
\mathbb E_{\mathcal C\sim\pi_\theta(\cdot\mid x)}
\mathbb E_{(y^+,y^-)\sim P(\cdot\mid x,\mathcal C)}\\
&\left[
E(x,y^+)-E(x,y^-)
\right]
\\
=&
\Delta_{\mathrm{gap}}(E;\theta)
\qquad\text{(by Equation~\eqref{eq:delta-gap})}.
\end{split}
\end{equation}
Therefore, $\Delta_{\mathrm{gap}}(E;\theta)$ is the expected local optimization signal exposed by the evaluator.
Evaluator updates should preserve these relative score gaps as optimization signals under policy-induced response set distribution shift.

\subsection{Response-Set-Conditioned Evaluation}
\label{sec:set-conditioned-eval}

Preserving this signal requires relative scores that adapt to the candidates being compared.
We therefore study how the evaluator input structure affects its capacity to model response-set-specific score gaps.
We compare two evaluator function classes.
A prompt-only evaluator assigns a fixed score $E(x,y)$ to a prompt-response pair, while a response-set-conditioned evaluator additionally takes the current response set as input and assigns a score $E(x,y\mid\mathcal C)$. Formally, consider
\begin{equation}
\small
\label{eq:evaluator-classes}
\begin{aligned}
\mathcal E_{\mathrm{prompt}}
&=
\{E:(x,y)\mapsto E(x,y)\in\mathbb R\},
\\
\mathcal E_{\mathrm{set}}
&=
\{E:(x,y,\mathcal C)\mapsto E(x,y\mid\mathcal C)\in\mathbb R\}.
\end{aligned}
\end{equation}
A prompt-only evaluator is the special case satisfying $E(x,y\mid\mathcal C)=E(x,y)$.
Therefore,
$\mathcal E_{\mathrm{prompt}}
\subseteq
\mathcal E_{\mathrm{set}}$.
We define the local ranking loss functional
\begin{equation}
\small
\label{eq:local-ranking-loss}
\begin{split}
&\mathcal L_{x,\mathcal C}(E)\\
=&
\mathbb E_{(y^+,y^-)\sim P(\cdot\mid x,\mathcal C)}
\left[
\ell\left(
E(x,y^+\mid\mathcal C)-E(x,y^-\mid\mathcal C)
\right)
\right], 
\end{split}
\end{equation}
where $\ell$ is any fixed ranking loss that is monotone non-increasing in the relative score gap.
By the function-class inclusion, we have
\begin{equation}
\small
\label{eq:set-evaluator-better-inf}
\begin{split}
\inf_{E\in\mathcal E_{\mathrm{prompt}}}&
\mathbb E_{\mathcal C\sim\pi_\theta(\cdot\mid x)}
\left[
\mathcal L_{x,\mathcal C}(E)
\right]\\
\ge
\inf_{E\in\mathcal E_{\mathrm{set}}}&
\mathbb E_{\mathcal C\sim\pi_\theta(\cdot\mid x)}
\left[
\mathcal L_{x,\mathcal C}(E)
\right]. 
\end{split}
\end{equation}
Equation~\ref{eq:set-evaluator-better-inf} shows that a response-set-conditioned evaluator has no larger optimal expected local ranking loss than a prompt-only evaluator.
This motivates using response-set-conditioned evaluators as the basis for preserving relative score gaps.

\subsection{Bilevel Policy-Evaluator Updates under Policy-Induced Response-Set Shift}
\label{sec:bilevel}

The remaining issue is training-time response-set shift, since policy optimization changes the response sets on which relative gaps must be preserved.
As policy optimization improves generations, candidate responses become closer in quality.
Reward models and LLM judges are known to be less reliable on such fine-grained comparisons~\citep{tan2025judgebench,lambert2025rewardbench,li2024crowdsourced}.
Consequently, an evaluator trained on earlier response sets may provide weaker relative signals on current policy samples.
Therefore, the evaluator must be updated along with policy-induced response sets to maintain the relative scoring structure required for policy optimization.

This joint update requirement naturally yields a coupled optimization problem.
The policy update depends on response-level scores produced by the current evaluator, while the evaluator update depends on response sets sampled from the current policy.
For evaluator $\phi$ and policy $\theta$ at round $t$, we formalize this as the following bilevel problem:
\begin{equation}
\small
\label{eq:bilevel}
\begin{aligned}
\max_{\theta}\quad
& \mathcal J_{\mathrm{policy}}^t(\theta;\phi_t^\star)
\\
\text{s.t.}\quad
& \phi_t^\star
\in
\arg\max_{\phi}\;
\mathcal J_{\mathrm{evaluator}}^t(\phi;\theta_{t-1}).
\end{aligned}
\end{equation}
This coupling makes evaluator learning track response-set shift during optimization and preserve relative discriminability for policy optimization.

\subsection{DynamicRubric}
\label{sec:dynamic-rubrics}

We instantiate this bilevel formulation with \textbf{DynamicRubric}, a co-evolution framework realized through a two-stage optimization loop.
At each round, DynamicRubric first fixes the current policy and adapts the evaluator on response sets sampled from that policy; it then fixes the updated evaluator and optimizes the policy using the resulting scores.
As illustrated in Figure~\ref{fig:main}, the evaluator consists of a learnable \textbf{DR-Generator} and a frozen \textbf{DR-Verifier}.
Given a prompt and a response set sampled from \textbf{DR-Policy}, DR-Generator produces weighted rubric items that define evaluation criteria and their relative importance.
DR-Verifier applies these rubrics to each response and returns binary outcomes for each criterion.
The weighted outcomes are aggregated into response-level scores.

Formally, given $(x,\mathcal C)$, DR-Generator outputs a rubric list $\{(r_m,w_m)\}_{m=1}^{M}$.
Given a response $y$, DR-Verifier returns a yes/no judgment for each rubric item, denoted by
$v_m(y;\mathcal C)\in\{0,1\}$.
Here, each $r_m$ is a rubric item and $w_m\in\{1,\cdots,5\}$ is its weight. 
We do not fix $M$ or prescribe item weights, allowing DR-Generator to determine both during training.
Aggregating rubric-level outcomes by normalized weights gives the response-level evaluator

\begin{equation}
\small
\label{eq:rubric-evaluator}
E_\phi(x,y\mid\mathcal C)
=
\frac{\sum_{m=1}^{M}w_m v_m(y;\mathcal C)}
{\sum_{m=1}^{M}w_m}.
\end{equation}

At round $t$, DynamicRubric fixes the current policy $\theta_{t-1}$ and samples a response set
$\mathcal C=\{y_i\}_{i=1}^{K}\sim\pi_{\theta_{t-1}}(\cdot\mid x)$.
These online sampled responses typically lack ranking labels.
Thus, on the current policy-induced distribution, the evaluator can be updated by a direction-free signal that encourages generated rubrics to separate responses.
For each rubric item $r_m$, define
$\bar v_m=\frac{1}{K}\sum_{i=1}^{K}v_m(y_i;\mathcal C)$ and the discriminability reward as
\begin{equation}
\small
\label{eq:disc-reward}
R_{\mathrm{disc}}(x,\mathcal C;\phi)
=
\frac{
\sum_{m=1}^{M}
w_m\,
\bar v_m(1-\bar v_m)
}{
\sum_{m=1}^{M}w_m
}.
\end{equation}
This reward is the weighted Bernoulli variance of rubric outcomes on $\mathcal C$.
It encourages DR-Generator to produce rubric items that induce non-trivial yes/no splits over the current response set.
The corresponding \textbf{discriminability objective} is
\begin{equation}
\small
\label{eq:j-disc}
\mathcal J_{\mathrm{disc}}^{t}(\phi;\theta_{t-1})
=
\mathbb E_{x\sim\mathcal D}
\mathbb E_{\mathcal C\sim\pi_{\theta_{t-1}}(\cdot\mid x)}
\left[
R_{\mathrm{disc}}(x,\mathcal C;\phi)
\right].
\end{equation}
Optimizing this objective alone encourages separations, but does not determine whether they are quality-aligned.
Indeed, flipping all yes/no outcomes preserves $R_{\mathrm{disc}}$ while reversing the induced relative score gaps.
Thus, the objective can also reward superficial splits.
Updating the evaluator therefore requires a direction-aware ranking signal to calibrate which separations are useful.

Let $\mathcal A$ denote ranked anchor data.
Each anchor example contains a prompt $x$ and a fully ranked anchor response list
$\mathcal C_{\mathrm{anchor}}=(y_1^0,\dots,y_n^0)$ satisfying $y_1^0\succ y_2^0\succ\cdots\succ y_n^0$.
Given a conditioning response set $\mathcal C$, DR-Generator observes only $(x,\mathcal C)$, and the rubrics generated from $\mathcal C$ are applied to $\mathcal C_{\mathrm{anchor}}$.
The induced score gaps provide preference-alignment supervision, calibrating local separations without exposing anchor-specific shortcuts.
For each anchor response $y_i^0\in\mathcal C_{\mathrm{anchor}}$, define $S_i=E_\phi(x,y_i^0\mid\mathcal C)$.
The anchor reward uses a Bradley--Terry style~\citep{bradley1952rank} contrastive ranking objective with Rank-Biased Precision style~\citep{moffat2008rank} weighting:
\begin{equation}
\small
\label{eq:anchor-reward}
R_{\mathrm{anchor}}(x,\mathcal C,\mathcal C_{\mathrm{anchor}};\phi)
=
\sum_{i=1}^{n-1}\sum_{j=i+1}^{n}
\frac{\log\sigma(S_i-S_j)}{(j-i)\times2^{i-1}}.
\end{equation}
Here, $\sigma(\cdot)$ is the logistic sigmoid.
The term $\log\sigma(S_i-S_j)$ encourages score gaps in the anchor ranking direction.
The weighting scheme increases the weight on top-ranked responses and emphasizes fine-grained distinctions between nearby ranked responses.
The corresponding \textbf{anchor objective} is
\begin{equation}
\small
\label{eq:j-anchor}
\begin{split}
&\mathcal J_{\mathrm{anchor}}^{t}(\phi;\theta_{t-1})
\\&=
\mathbb E_{(x,\mathcal C_{\mathrm{anchor}})\sim\mathcal A}
\mathbb E_{\mathcal C\sim\pi_{\theta_{t-1}}(\cdot\mid x)}
\left[
R_{\mathrm{anchor}}(x,\mathcal C,\mathcal C_{\mathrm{anchor}};\phi)
\right].
\end{split}
\end{equation}
Combining these two objectives gives the \textbf{evaluator objective}
\begin{equation}
\small
\label{eq:j-evaluator}
\mathcal J_{\mathrm{evaluator}}^{t}(\phi;\theta_{t-1})
=
\mathcal J_{\mathrm{disc}}^{t}(\phi;\theta_{t-1})
+
\lambda
\mathcal J_{\mathrm{anchor}}^{t}(\phi;\theta_{t-1}),
\end{equation}
where $\lambda$ controls the relative scale of the anchor reward.
Although $R_{\mathrm{anchor}}$ is non-positive, maximizing it directly improves agreement with the anchor ranking.
The evaluator objective is designed as a tractable proxy for the \textbf{relative score gap $\Delta_{\mathrm{gap}}(E_\phi;\theta_{t-1})$ in Equation~\ref{eq:gap-as-expected-directional-gain}}.
Appendix~\ref{app:alignment-guarantee} further shows that there exist constants
$c_{\mathrm{disc}},c_{\mathrm{anchor}}>0$ and a $t$-dependent residual $\epsilon_t$ such that
\begin{equation}
\small
\label{eq:main-gap-lower-bound}
\begin{split}
&\Delta_{\mathrm{gap}}(E_\phi;\theta_{t-1})
\\&\ge
c_{\mathrm{disc}}\,
\mathcal J_{\mathrm{disc}}^{t}(\phi;\theta_{t-1})
+
c_{\mathrm{anchor}}\,
\mathcal J_{\mathrm{anchor}}^{t}(\phi;\theta_{t-1})
+
\epsilon_t.
\end{split}
\end{equation}
Therefore, maximizing the combination of the discriminability and anchor objectives increases a lower bound on the target relative score gap.

\begin{table*}[t]
\centering
\small
\setlength{\tabcolsep}{4.2pt}
\renewcommand{\arraystretch}{1.12}
\resizebox{\textwidth}{!}{
\begin{tabular}{llcccccccc}
\toprule[1.2pt]
\multirow{2}{*}{\textbf{Evaluator}}
& \multirow{2}{*}{\textbf{Size}}
& \textbf{JudgeBench}
& \textbf{Personalized-RB}
& \textbf{RM-Bench}
& \textbf{UltraFeedback}
& \multicolumn{2}{c}{\textbf{AEOLLM}}
& \multicolumn{2}{c}{\textbf{LMSYS-Chat-1M}} \\
&
& \multicolumn{1}{c}{Acc.$\uparrow$}
& \multicolumn{1}{c}{Acc.$\uparrow$}
& \multicolumn{1}{c}{Acc.$\uparrow$}
& \multicolumn{1}{c}{Acc.$\uparrow$}
& \multicolumn{1}{c}{Acc.$\uparrow$}
& \multicolumn{1}{c}{nDCG$\uparrow$}
& \makebox[3em][c]{Acc.$\uparrow$}
& \makebox[3.2em][c]{nDCG$\uparrow$} \\
\midrule[0.8pt]

\multicolumn{10}{l}{\textbf{Scalar Reward Models}} \\
\addlinespace[1pt]
Skywork-27B
& 27B
& 52.1
& 64.7
& 67.7
& 63.4
& 43.8
& 0.826
& 21.0
& 0.720 \\

Skywork-27B w/ responses
& 27B
& 56.1
& 60.6
& 65.8
& 65.2
& 37.5
& 0.805
& 21.6
& 0.717 \\

Nemotron-70B
& 70B
& \textbf{66.0}
& 68.6
& \textbf{85.2}
& 72.8
& \textbf{57.5}
& 0.868
& 20.4
& 0.718 \\

Nemotron-70B w/ responses
& 70B
& 64.8
& 61.9
& 83.2
& \textbf{77.4}
& \textbf{57.5}
& 0.867
& 17.2
& 0.706 \\

\midrule[0.6pt]
\multicolumn{10}{l}{\textbf{Dynamic Rubric Methods}} \\
\addlinespace[1pt]
Qwen3-8B zero-shot
& 8B
& 36.4
& 53.7
& 52.2
& 54.7
& 17.8
& 0.894
& 12.0
& 0.762 \\

Qwen3-32B zero-shot
& 32B
& 46.4
& 67.6
& 64.7
& 64.6
& 29.7
& \textbf{\underline{0.902}}
& 15.1
& 0.763 \\

\rowcolor{mylightblue}
\textbf{DR-Generator-8B}
& 8B
& \underline{57.4}
& \textbf{\underline{73.3}}
& \underline{69.8}
& \underline{68.7}
& \underline{51.2}
& 0.898
& \textbf{\underline{28.0}}
& \textbf{\underline{0.779}} \\

\bottomrule[1.2pt]
\end{tabular}
}
\caption{
Evaluator performance on pairwise and listwise benchmarks.
Scalar reward models marked as ``w/ responses'' receive the candidate response set during scoring.
\textbf{DR-Generator-8B} uses Qwen3-8B as the backbone.
Acc. denotes strict (ties excluded) accuracy.
nDCG is reported for listwise benchmarks.
The overall best is in \textbf{bold} and the best among dynamic rubric generators under the same verifier (details in Appendix~\ref{app:verifier-robustness}) is \underline{underlined}.
}

\label{tab:evaluator-results}
\end{table*}

After the evaluator update, the \textbf{policy objective} uses the response-level reward from the evaluator:
\begin{equation}
\small
\label{eq:j-policy}
\mathcal J_{\mathrm{policy}}^t(\theta;\phi_t)
=
\mathbb E_{x\sim\mathcal D}
\mathbb E_{\mathcal C\sim\pi_\theta(\cdot\mid x)}
\left[
\frac{1}{|\mathcal C|}
\sum_{y\in\mathcal C}
E_{\phi_t}(x,y\mid\mathcal C)
\right].
\end{equation}
Thus, the \textbf{overall training procedure} is
\begin{equation}
\small
\label{eq:training-procedure}
\left\{
\begin{aligned}
\phi_t
&\in
\arg\max_{\phi}\;
\mathcal J_{\mathrm{evaluator}}^t(\phi;\theta_{t-1}),
\qquad t\ge 1,
\\
\theta_t
&\in
\arg\max_{\theta}\;
\mathcal J_{\mathrm{policy}}^t(\theta;\phi_t),
\qquad t\ge 1,
\end{aligned}
\right.
\end{equation}
where $\theta_0$ denotes the initial policy model. 

Notably, DynamicRubric can be implemented in a fully self-contained manner: given only ranked anchor data, DR-Generator, DR-Verifier, and DR-Policy can be instantiated as separate copies of the same backbone model, without requiring any external reward model or judge.

\begin{table*}[t]
\centering
\small
\setlength{\tabcolsep}{6.0pt}
\renewcommand{\arraystretch}{1.12}
\resizebox{\textwidth}{!}{
\begin{tabular}{lclcccc}
\toprule[1.2pt]
\textbf{Supervision Method}
& \textbf{Supervisor Size}
& \textbf{Backbone}
& \textbf{AlpacaEval2}
& \textbf{ArenaHardv2.0}
& \textbf{WildBench}
& \textbf{WritingBench} \\
\midrule[0.8pt]

\multicolumn{7}{l}{\textbf{Baselines}} \\
\addlinespace[1pt]
Native
& --
& Qwen3-8B
& 38.5
& 9.0
& 16.4
& 57.8 \\

Native
& --
& Qwen3-32B
& 50.2
& 20.8
& 29.7
& 62.0 \\

\midrule[0.6pt]
\multicolumn{7}{l}{\textbf{Scalar Reward Models}} \\
\addlinespace[1pt]
Skywork-27B
& 27B
& Qwen3-8B
& 50.6
& 15.3
& 22.3
& 60.2 \\

Skywork-27B w/ responses
& 27B
& Qwen3-8B
& 48.2
& 12.3
& 20.5
& 58.6 \\

Nemotron-70B
& 70B
& Qwen3-8B
& 35.2
& 10.3
& 17.5
& 56.2 \\

Nemotron-70B w/ responses
& 70B
& Qwen3-8B
& 41.9
& 9.2
& 18.9
& 58.2 \\

\midrule[0.6pt]
\multicolumn{7}{l}{\textbf{Static Rubric}} \\
\addlinespace[1pt]
RaR~\citep{gunjal2025rubrics}
& 235B
& Qwen3-8B
& 58.5
& 15.7
& 25.2
& 61.6 \\

\midrule[0.6pt]
\multicolumn{7}{l}{\textbf{Dynamic Rubric}} \\
\addlinespace[1pt]
\rowcolor{mylightblue}
\textbf{DR-Generator-8B}
& 8B
& Qwen3-8B
& \textbf{67.1}
& \textbf{21.0}
& \textbf{30.7}
& \textbf{64.3} \\

\bottomrule[1.2pt]
\end{tabular}
}
\caption{
Policy performance on open-domain generation benchmarks across different supervision methods.
AlpacaEval2 and ArenaHardv2.0 report win rates.
WildBench and WritingBench report benchmark scores in their standard evaluation scales.
All scores are reported such that higher is better.
In each column, the best is in \textbf{bold}.
}
\label{tab:policy-results}
\end{table*}

\begin{figure*}[t]
    \centering
    \begin{subfigure}{0.95\linewidth}
        \centering
        \includegraphics[width=\linewidth]{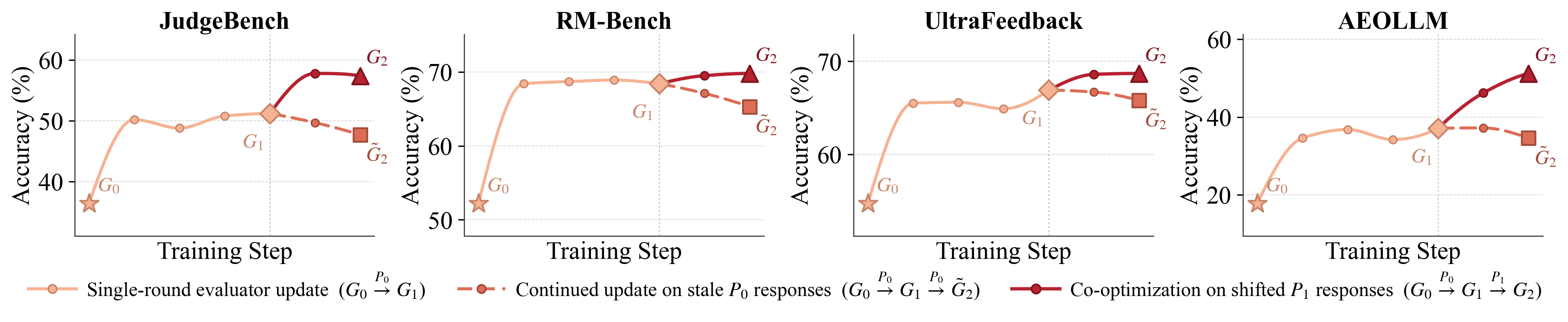}
        \caption{DR-Generator optimization on original versus updated DR-Policy response distribution.}
        \label{fig:continuous-generator}
    \end{subfigure}

    \vspace{0.6em}

    \begin{subfigure}{0.95\linewidth}
        \centering
        \includegraphics[width=\linewidth]{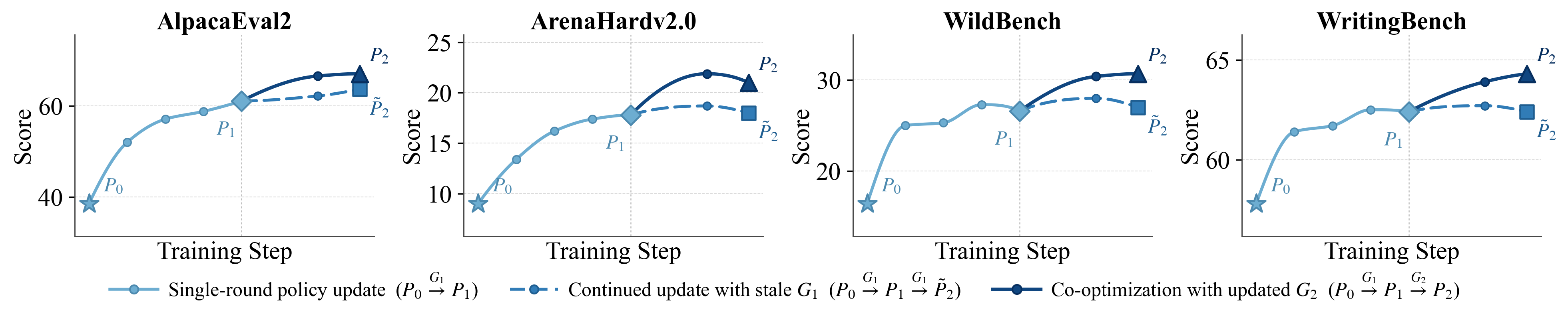}
        \caption{DR-Policy optimization with dynamic rubric supervision from previous versus updated DR-Generators.}
        \label{fig:continuous-policy}
    \end{subfigure}

    \caption{
    Effect of evaluator--policy co-evolution in DynamicRubric.
$G_0$ and $P_0$ are separately initialized from Qwen3-8B.
The co-evolution process starts by adapting $G_0$ on responses sampled from $P_0$.
    }
    \label{fig:continuous}
\end{figure*}

\section{Experiments}
\subsection{Experimental Setup}
\label{sec:exp-setup}

\paragraph{Datasets.}
We construct the training data from Nectar~\citep{starling2023}, an open-domain dataset with seven ranked responses for each prompt, which we use as the anchor data.
For evaluator evaluation, we use four pairwise and two listwise preference benchmarks and report AVG@5. For policy evaluation, we use four open-ended generation benchmarks. Details of the datasets and benchmarks are provided in Appendices~\ref{app:dataset-details} and~\ref{app:benchmark-details}.

\paragraph{Models.}
We initialize DR-Policy and DR-Generator from independent Qwen3-8B~\citep{yang2025qwen3} copies and optimize both with GRPO~\citep{shao2024deepseekmath}.
For DR-Policy, GRPO turns evaluator scores into relative advantages within each sampled set, aligning policy optimization with the score-gap view.
For DR-Generator, discriminability and anchor rewards are combined at the advantage level, which reduces scale mismatch and allows us to set $\lambda=1$.
Training details are provided in Appendix~\ref{app:training-details}.
We repeat the full pipeline with Llama-3.1-8B-Instruct~\citep{grattafiori2024llama3herdmodels} in Appendix~\ref{app:additional-experiments-llama}, demonstrating robustness to model backbones.

\subsection{Evaluator Optimization with DynamicRubric}
\label{sec:evaluator-performance}

We first examine whether DynamicRubric improves fine-grained preference evaluation.
Our evaluator uses a DR-Generator initialized from Qwen3-8B and trained with DynamicRubric, denoted \textbf{DR-Generator-8B}.
We compare it with the larger same-family Qwen3-32B used as a dynamic rubric generator, and with strong scalar reward models, including Skywork-Reward-Gemma2-27B~\citep{liu2024skywork} and Nemotron-70B-Reward-Multilingual~\citep{wang2025helpsteer3preferenceopenhumanannotatedpreference}.
To control for response-set access, we also evaluate scalar reward models that receive the same response set during scoring.
This setup isolates the effect of rubric generator training under matched response-set access while comparing against strong baselines.

Table~\ref{tab:evaluator-results} shows that DR-Generator-8B consistently improves over zero-shot dynamic rubric evaluators, including the larger Qwen3-32B generator, on all accuracy metrics.
In contrast, the scalar variants show mixed changes, indicating that response-set access alone does not reliably improve evaluation.
Notably, DR-Generator-8B also outperforms Skywork-27B and its response-set variant on every reported metric.
Overall, DR-Generator-8B provides consistent gains over larger zero-shot dynamic rubric evaluators and achieves performance competitive with substantially larger scalar reward models.
We also include qualitative analysis and a case study in Appendices~\ref{app:rubric-quality} and~\ref{app:case-study} to examine whether DynamicRubric training turns response-set access into high-quality rubrics that identify decisive differences.
The analysis shows that training yields more discriminative criteria, and the case study illustrates how these criteria produce clear score gaps and the correct top-ranked response.

\subsection{Policy Optimization with DynamicRubric}
\label{sec:policy-optimization}

Next, we evaluate whether these finer score gaps translate into effective downstream DR-Policy supervision signals.
We compare against other supervision approaches including scalar reward models, their response-set-conditioned variants, and Rubric-as-Reward~\citep{gunjal2025rubrics}, a static rubric supervision approach where rubrics are generated only from the prompt with Qwen3-235B-A22B-Instruct-2507 as the generator.
Qwen3-32B is also included as a same-family strong policy baseline.

Table~\ref{tab:policy-results} shows that DR-Generator-8B supervision produces a DR-Policy that consistently outperforms policies trained with larger reward models, static rubric supervision from a 235B generator, and the strong 32B baseline across all benchmarks.
We use DeepSeek-V4-Flash~\citep{deepseekai2026deepseekv4} as the judge. Appendix~\ref{app:judge-robustness} re-evaluates policies with GPT-4.1 and reports consistent gains, supporting the reliability of the evaluation.
Together, these results show that DynamicRubric converts response-set-conditioned evaluator gains into effective reward signals for policy optimization.

\begin{table*}[t]
\centering
\small
\setlength{\tabcolsep}{4.8pt}
\renewcommand{\arraystretch}{1.12}
\resizebox{\textwidth}{!}{
\begin{tabular}{lcc cccccc c}
\toprule[1.2pt]
\multirow{2}{*}{\textbf{Evaluator Objective}}
& \multirow{2}{*}{$\mathcal J_{\mathrm{disc}}$}
& \multirow{2}{*}{$\mathcal J_{\mathrm{anchor}}$}
& \textbf{JudgeBench}
& \textbf{Personalized-RB}
& \textbf{RM-Bench}
& \textbf{UltraFeedback}
& \textbf{AEOLLM}
& \textbf{LMSYS-Chat-1M}
& \textbf{Average} \\
&
&
& Acc.$\uparrow$
& Acc.$\uparrow$
& Acc.$\uparrow$
& Acc.$\uparrow$
& Acc.$\uparrow$
& Acc.$\uparrow$
& Macro$\uparrow$ \\
\midrule[0.8pt]

w/o anchor objective
& \checkmark
& -
& 56.1
& 72.4
& \textbf{69.8}
& 67.9
& 47.3
& 27.7
& 56.9 \\

w/o discriminability objective
& -
& \checkmark
& 50.0
& \textbf{73.8}
& 66.3
& 66.8
& 38.8
& 20.7
& 52.7 \\

\rowcolor{mylightblue}
\textbf{Full objective}
& \checkmark
& \checkmark
& \textbf{57.4}
& 73.3
& \textbf{69.8}
& \textbf{68.7}
& \textbf{51.2}
& \textbf{28.0}
& \textbf{58.1} \\

\bottomrule[1.2pt]
\end{tabular}
}
\caption{
Ablation of evaluator optimization objectives.
In each column, the best is in \textbf{bold}.
}
\label{tab:evaluator-objective-ablation}
\end{table*}

\subsection{Continuous Co-Evolution of Evaluator and Policy}
\label{sec:continuous-cooptimization}

We then test the continuous co-evolution claim by isolating evaluator and policy updates.
Let $G_0$ and $P_0$ be two separate copies initialized from the native Qwen3-8B backbone.
The first DR-Generator update, $G_0 \overset{P_0}{\rightarrow} G_1$, trains the generator to produce dynamic rubrics for response sets sampled from $P_0$.
The resulting $G_1$ then supervises DR-Policy optimization, $P_0 \overset{G_1}{\rightarrow} P_1$, shifting the policy-induced response distribution.
The next round isolates two dependencies in this co-evolution loop.
For evaluator training, we compare continuing to train on the original policy distribution, $G_1 \overset{P_0}{\rightarrow} \tilde{G}_2$, against updating on the current policy distribution, $G_1 \overset{P_1}{\rightarrow} G_2$.
For policy optimization, we compare updating $P_1$ with the previous DR-Generator, $P_1 \overset{G_1}{\rightarrow} \widetilde{P}_2$, against updating it with the current DR-Generator, $P_1 \overset{G_2}{\rightarrow} P_2$.

Figure~\ref{fig:continuous} shows the empirical effect of continuous evaluator--policy co-evolution in DynamicRubric.
For evaluator adaptation, training DR-Generator on responses from the current DR-Policy yields stronger gains, while continuing on the stale $P_0$ distribution can degrade.
For policy optimization, the updated generator $G_2$ provides stronger supervision for $P_1$ and improves all four generation benchmarks.
In contrast, the stale generator $G_1$ yields smaller gains, as criteria calibrated to $P_0$ can saturate or induce noisy gaps on $P_1$ samples.
These results support the core design principle of DynamicRubric: evaluator learning should track the evolving policy-induced response distribution, and policy optimization should use the evaluator most recently adapted to that distribution.

\begin{figure}[t]
    \centering
    \includegraphics[width=\linewidth]{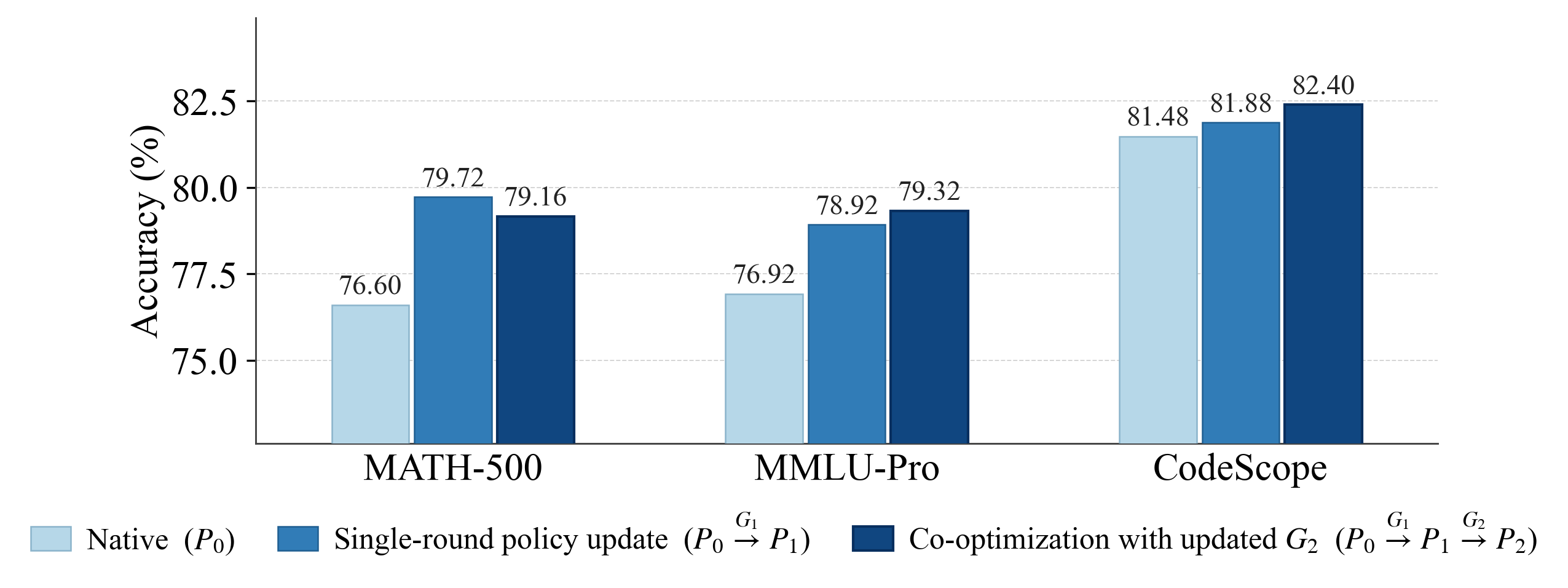}
    \caption{
    Performance of the initial policy and two DynamicRubric optimized policies on verifiable tasks.
    }
    \label{fig:general-bench}
\end{figure}

\subsection{Online A/B Test and Deployment in Production}
\label{sec:online-ab}

Finally, we evaluate DynamicRubric in a large-scale production environment: WeChat Search's AI answering scenario, which serves tens of millions of requests per day.
We use the previous production model as the online baseline.
Both models use \textbf{WeLM-V4-80B-A3B}~\cite{wechat_ai_team_2026_sparse_moe}, an large language model developed by \textbf{WeChat}.
Due to business confidentiality, we omit the exact values of online measurements and describe the production results textually.
In the online A/B test, the DynamicRubric-optimized model achieves statistically significant relative improvements over the production baseline in total search volume, user duration, and absolute positive user behaviors.
Given the stable online gains, the resulting model is fully deployed in production and currently serves all online traffic, replacing the previous production model.
These results demonstrate that the policy supervision produced by DynamicRubric remains effective at production scale in a real-world AI answering scenario.
In production, DynamicRubric further enables anchor-guided continual self-improvement: refreshed anchors recalibrate the direction of evaluator feedback, allowing subsequent policy iterations to track evolving objectives.

\section{Analysis}
\subsection{Ablation Study on Evaluator Optimization Objectives}
\label{sec:evaluator-objective-ablation}

We ablate two objective terms used to optimize the evaluator in DynamicRubric: the discriminability objective $\mathcal J_{\mathrm{disc}}$ and the anchor objective $\mathcal J_{\mathrm{anchor}}$.
All variants start from the same DR-Generator and are optimized on response sets sampled from the same policy for the same number of steps, isolating the effect of the objective components.

Table~\ref{tab:evaluator-objective-ablation} shows that the full objective achieves the highest macro average across benchmarks.
Removing $\mathcal J_{\mathrm{anchor}}$ leaves the signal vulnerable to separations that are not quality-relevant, while removing $\mathcal J_{\mathrm{disc}}$ weakens adaptation to current response sets.
This pattern indicates that separation and ranking calibration provide complementary supervision for DR-Generator, and their combination yields a stronger evaluator across benchmarks.

\subsection{DynamicRubric Policy Gains Transfer to Verifiable Tasks}
\label{sec:verifiable-tasks}

DR-Policies are optimized for open-ended generation.
Prior work reports that open-ended generation optimization can cause regressions on reasoning and coding tasks~\citep{ouyang2022training,kirk2024understandingeffectsrlhfllm}.
This raises the question of whether the benefits of DynamicRubric extend to reasoning abilities.
We evaluate DR-Policies and report AVG@5 on mathematics, scientific QA, and coding benchmarks detailed in Appendix~\ref{app:benchmark-details}.
Figure~\ref{fig:general-bench} shows that DR-Policies consistently improve over the initial policy across these out-of-distribution tasks, indicating that DynamicRubric improves open-ended generation and that its policy-level benefits transfer to verifiable reasoning and coding tasks.

\section{Conclusion}

We identify a central bottleneck in evaluator-guided policy optimization: as current-policy responses become close in quality, effective updates require evaluators to maintain discriminability.
We formalize this requirement through a probability allocation view, where moving probability mass between two responses yields a gain equal to their evaluator score gap.
This yields a simple principle: evaluators should evolve with the policies they supervise, adapting to current-policy response sets before the next policy update.
We propose \textbf{DynamicRubric} to instantiate this principle with response-set-conditioned rubrics aggregated into response-level scores for supervision.
Experiments show gains in evaluation, open-ended generation, and verifiable tasks, highlighting relative discriminability on the policy-induced response distribution as central to effective post-training evaluators.

\section*{Limitations}

\paragraph{Training-time computational cost.}
DynamicRubric introduces additional training-time cost because policy optimization requires generator and verifier calls.
Although Appendix~\ref{app:computational-overhead} reports that this overhead remains lower than supervision from larger reward model or rubric generator baselines in our setup, serial generator--verifier calls still increase wall-clock latency during policy training.
This cost may become more significant with larger generators, stronger verifiers, or larger response sets.
Future work can reduce this overhead through parallel verification, asynchronous evaluator updates, and more efficient verifier scheduling.

\paragraph{Verifier scope.}
DynamicRubric relies on verifier feedback to score responses under generated rubrics, so the verifier choice defines the supervision available during training.
Although Appendix~\ref{app:verifier-robustness} evaluates verifiers of different scales and from different model families, our study does not include stronger black-box verifiers such as GPT-4.1 during training because of evaluation cost.
We also keep DR-Verifier frozen rather than training a dedicated verifier.
Future work can study stronger verifier families, adaptive verifier training, and calibration strategies for rubric-based scoring.

\bibliography{custom}
\appendix
\clearpage

\begin{algorithm*}[t]
\caption{DynamicRubric Training}
\label{alg:dynamicrubric}
\begin{algorithmic}[1]
\Require Initial policy $P_0=\pi_{\theta_0}$; initial generator $G_0=G_{\phi_0}$; frozen verifier $V$; prompt distribution $\mathcal D$; ranked anchor data $\mathcal A$; rounds $T$; response-set size $K$; anchor weight $\lambda$.
\Ensure Trained generator $G_T$ and trained policy $P_T$.

\For{$t=1,\dots,T$}
    \Statex \textbf{Evaluator update: $G_{t-1}\rightarrow G_t$ using samples from $P_{t-1}$.}
    \State Initialize $\phi\leftarrow\phi_{t-1}$.
    \For{generator update steps}
        \State Sample $(x,\mathcal C_{\mathrm{anchor}})\sim\mathcal A$ and current-policy responses $\mathcal C=\{y_i\}_{i=1}^{K}$ with $y_i\sim P_{t-1}(\cdot\mid x)$.
        \State Generate rubrics $\mathcal R_{\phi}(x,\mathcal C)=\{(r_m,w_m)\}_{m=1}^{M}$.
        \State Use $V$ to verify responses in $\mathcal C$ and compute $R_{\mathrm{disc}}(x,\mathcal C;\phi)$.
        \State Use the same rubrics to verify $\mathcal C_{\mathrm{anchor}}$ and compute $R_{\mathrm{anchor}}(x,\mathcal C,\mathcal C_{\mathrm{anchor}};\phi)$.
        \State Update $\phi$ to maximize $R_{\mathrm{disc}}+\lambda R_{\mathrm{anchor}}$.
    \EndFor
    \State Set $\phi_t\leftarrow\phi$ and $G_t\leftarrow G_{\phi_t}$.

    \Statex \textbf{Policy update: $P_{t-1}\rightarrow P_t$ using the fixed generator $G_t$.}
    \State Initialize $\theta\leftarrow\theta_{t-1}$.
    \For{policy update steps}
        \State Sample $x\sim\mathcal D$ and responses $\mathcal C=\{y_i\}_{i=1}^{K}$ with $y_i\sim\pi_{\theta}(\cdot\mid x)$.
        \State Generate rubrics $\mathcal R_{\phi_t}(x,\mathcal C)$ using fixed $G_t$.
        \State Use $V$ to compute scores $s_i=E_{\phi_t}(x,y_i\mid\mathcal C)$.
        \State Update $\theta$ with the policy optimizer using the rubric-based scores.
    \EndFor
    \State Set $\theta_t\leftarrow\theta$ and $P_t\leftarrow\pi_{\theta_t}$.
\EndFor

\State \Return $G_T$ and $P_T$.
\end{algorithmic}
\end{algorithm*}

\section{DynamicRubric Training Algorithm}
\label{app:dynamicrubric-algorithm}

Algorithm~\ref{alg:dynamicrubric} summarizes DynamicRubric training.
The first sampled response sets come from the initial policy $P_0$, so the first update adapts $G_0$ to $G_1$.
The updated generator $G_1$ then supervises the first policy update from $P_0$ to $P_1$.

For GRPO, rubric scores are converted into group-normalized advantages, so the policy update depends on within-set relative scores rather than absolute reward scale.

\section{Dataset Details}
\label{app:dataset-details}

\subsection{Dataset Details for Training DR-Generator}

For DR-Generator optimization, each training instance contains a user prompt and a candidate response set.
We apply lightweight filtering to ensure that the generator observes sufficiently informative inputs while avoiding response sets dominated by degenerate model outputs.

We first apply length-based filtering.
Each example must contain at least seven candidate responses.
The user prompt and the top-ranked response must each contain more than 50 characters.
We then apply response-set-level filtering to the first seven responses.
A response set is discarded if it contains any repetitive generation, if at least four responses are degenerate, if fewer than four responses are useful, or if at least four responses consist mostly of boilerplate.
Here, degenerate responses include broken text, plain refusals, greeting-only replies, and clarification-only replies, while useful responses include substantive answers, refusals with helpful alternatives, and other nontrivial responses.
After filtering, we obtain 24K training samples for DR-Generator optimization.

\subsection{Dataset Details for Training DR-Policy}

For DR-Policy optimization, each training instance is a user prompt.
We apply lightweight filtering to retain prompts that are sufficiently informative and whose associated source responses are not degenerate.
Each example must contain at least seven associated responses.
For most sources, the user prompt and the top-ranked associated response must each contain more than 50 characters.
For short-prompt sources whose prompts are already filtered upstream, we exempt the prompt from the length constraint, while still requiring the top-ranked response to be longer than 50 characters.

We further discard examples whose top-ranked associated response contains broken text, is repetitive, is a plain refusal, or consists only of a greeting or clarification request.
This filtering ensures that DR-Policy is optimized on prompts for which meaningful response generation can be supervised by the current DR-Generator.

After filtering, we retain 36K prompts for DR-Policy optimization.

\section{Benchmark Details}
\label{app:benchmark-details}

We evaluate DynamicRubric along three axes.
First, evaluator evaluation measures whether the trained DR-Generator produces rubrics that induce accurate response rankings.
Second, open-ended generation evaluation measures whether DR-Policy improves general instruction-following and conversational quality.
Third, verifiable task evaluation measures whether optimization with open-ended dynamic rubric supervision preserves capabilities on reasoning-heavy out-of-distribution tasks.

\subsection{Benchmark Details for Evaluator Evaluation}

For evaluator evaluation, we use pairwise preference benchmarks including JudgeBench~\citep{tan2025judgebench}, Personalized-RewardBench~\citep{ma2026personalizedrewardbenchevaluatingreward}, RM-Bench~\citep{liu2025rm}, and UltraFeedback~\citep{cui2023ultrafeedback}, and listwise preference benchmarks including AEOLLM~\citep{chen2025overviewntcir18automaticevaluation} and LMSYS-Chat-1M~\citep{zheng2024lmsys}.
For pairwise benchmarks, the evaluator is asked to select the preferred response from a candidate pair.
For listwise benchmarks, the evaluator assigns scores to multiple candidate responses, and the induced ranking is compared with the annotated ranking.

When evaluating dynamic rubric generators in Section~\ref{sec:evaluator-performance}, we use Qwen3-30B-A3B as the default verifier for all dynamic rubric methods.
This choice keeps the verifier fixed while avoiding same-model coupling with the generator baselines, which include Qwen3-8B and Qwen3-32B.
Using a separate model as the verifier ensures that no compared generator is evaluated by a verifier initialized from the same model instance or scale.
Section~\ref{app:verifier-robustness} further evaluates the same generators under multiple verifier choices, including Llama-3.1-8B-Instruct and Qwen3 verifiers of different scales.
DR-Generator-8B consistently improves over the zero-shot dynamic rubric generators across these verifier choices, including the cross-family Llama verifier, indicating that its gains are not tied to a particular verifier model.

\paragraph{JudgeBench.}
JudgeBench is a preference benchmark for evaluating LLM-as-a-judge and reward model behavior.
We use it as a pairwise evaluator benchmark, where each example contains a prompt, two candidate responses, and an annotated preference.

\paragraph{Personalized-RewardBench.}
Personalized-RewardBench evaluates reward models under personalized scenarios.
We use it as a pairwise evaluator benchmark, where each example contains a prompt, two candidate responses, and an annotated preference.

\paragraph{RM-Bench.}
RM-Bench is a reward model evaluation benchmark built around response preference comparisons.
We use it as a pairwise evaluator benchmark, where each example contains a prompt, two candidate responses, and an annotated preference.

\paragraph{UltraFeedback.}
UltraFeedback contains prompts paired with multiple model responses and preference-style annotations.
We use it as a pairwise evaluator benchmark, where each example contains a prompt, two candidate responses, and an annotated preference.

\paragraph{AEOLLM.}
AEOLLM is a listwise evaluation benchmark for automatic LLM evaluation.
We use it to measure whether the evaluator can rank multiple candidate responses consistently with the annotated ordering.
We report both strict top-1 accuracy and nDCG~\citep{jarvelin2002cumulated}.

\paragraph{LMSYS-Chat-1M.}
LMSYS-Chat-1M contains real-world user conversations from large-scale chat deployments.
Nectar provides the responses and ranking annotations for this subset; we hold out the complete subset as the listwise preference benchmark, with evaluation conducted only on this subset and training data drawn exclusively from other Nectar subsets to avoid leakage.
We report both strict top-1 accuracy and nDCG.

\subsection{Benchmark Details for Open-Ended Generation Tasks Evaluation}

For policy evaluation, we use DeepSeek-V4-Flash~\citep{deepseekai2026deepseekv4} as the judge on four open-domain generation benchmarks: AlpacaEval2~\citep{alpaca_eval,dubois2023alpacafarm,dubois2024length}, ArenaHardv2.0~\citep{li2024crowdsourced}, WildBench~\citep{yuchen2024wildbench}, and WritingBench~\citep{wu2025writingbench}.
AlpacaEval2 and ArenaHardv2.0 report win rates, while WildBench and WritingBench report benchmark scores in their standard evaluation scales.
All metrics are reported such that higher is better.
To assess robustness to the choice of automatic judge, Appendix~\ref{app:judge-robustness} further re-evaluates the same policy checkpoints with GPT-4.1 and reports consistent gains.

\paragraph{AlpacaEval2.}
AlpacaEval2 evaluates instruction-following ability through model-based pairwise comparisons on open-ended user instructions.
We use it to measure whether DR-Policy improves general assistant-style response quality.

\paragraph{ArenaHardv2.0.}
ArenaHardv2.0 contains challenging open-ended prompts designed to stress-test instruction-following models.
We use it to evaluate policy performance on difficult conversational and reasoning-heavy user requests.

\paragraph{WildBench.}
WildBench evaluates models on real-world conversational tasks.
We use it to measure whether DR-Policy improves response quality under diverse user-facing interaction scenarios.

\paragraph{WritingBench.}
WritingBench evaluates long-form writing ability.
We use it to test whether DynamicRubric optimization improves policy behavior on tasks requiring coherent, structured, and high-quality written responses.

\subsection{Benchmark Details for Verifiable Tasks Evaluation}

For verifiable-task evaluation, we use MATH-500~\citep{lightman2023lets} for mathematical reasoning, MMLU-Pro~\citep{wang2024mmlu} for scientific question answering, and CodeScope~\citep{yan2024codescopeexecutionbasedmultilingualmultitask} for coding.
MATH-500 is evaluated with Math-Verify~\citep{Kydlicek_Math-Verify_Math_Verification}, while MMLU-Pro and CodeScope are evaluated by exact match.

\paragraph{MATH-500.}
MATH-500 is a mathematical reasoning benchmark consisting of competition-style math problems.
We use Math-Verify~\citep{Kydlicek_Math-Verify_Math_Verification} (it is Apache-2.0 licensed and we follow its instructions in the repository) to check whether the model's final answer matches the reference answer.

\paragraph{MMLU-Pro.}
MMLU-Pro is a challenging question-answering benchmark.
We use its scientific question-answering setting and evaluate predictions by exact match.

\paragraph{CodeScope.}
CodeScope is a coding benchmark for evaluating programming ability.
We evaluate model outputs by exact match.

\section{Training Details}
\label{app:training-details}

We train our models with GRPO~\citep{Guo_2025,shao2024deepseekmath} as implemented in WeChat-YATT~\citep{WeChatYATTScalable2025}.
We use a rollout size of $8$ samples per instance, corresponding to $K=8$ in Equation~\ref{eq:sample-response-set}, a global batch size of $256$, $\beta=0.01$, and learning rates of $5\times10^{-6}$ for DR-Generator and $1\times10^{-6}$ for DR-Policy.
The maximum input/output lengths are 8K/8K for DR-Policy and 24K/8K for DR-Generator, as DR-Generator scores the full response set from each rollout.
We do not constrain the number of generated rubrics $M$ or their weights, but let DR-Generator learn this allocation from data; empirically, it learns reasonable schemes as the experimental results show.

GRPO replaces the separate value function~\citep{schulman2017proximalpolicyoptimizationalgorithms,mnih2016asynchronousmethodsdeepreinforcement} with a group-based relative advantage estimation~\citep{ahmadian2024basicsrevisitingreinforcestyle}. For each question $q$, the policy $\pi_{\theta_{\mathrm{old}}}$ generates $G$ candidate outputs $\{o_i\}_{i=1}^G$. The advantage for each token $o_{i,t}$ is computed as
\begin{equation}
\label{eq:GRPOadvantage}
\hat{A}_{i,t} = \frac{R_i - \mathrm{mean}(\{R_j\}_{j=1}^G)}{\mathrm{std}(\{R_j\}_{j=1}^G)},
\end{equation}
where $R_i$ denotes the scalar reward assigned to output $o_i$. This formulation normalizes rewards within the group. In the native GRPO implementation, the reward is binary and determined by an automatic rule-based verifier:
\begin{equation}
\label{GRPOreward}
R_i =
\begin{cases}
1, & \text{if the verifier returns \texttt{true} for output } o_i, \\
0, & \text{otherwise}.
\end{cases}    
\end{equation}

\begin{table*}[t]
\centering
\setlength{\tabcolsep}{6.0pt}
\renewcommand{\arraystretch}{1.12}
\resizebox{\textwidth}{!}{
\begin{tabular}{lclccc}
\toprule[1.2pt]
\textbf{Supervision Method}
& \textbf{Supervisor Size}
& \textbf{Backbone}
& \textbf{Verifier Needed}
& \textbf{Minimal GPU Allocation}
& \textbf{Actual GPU Hours} \\
\midrule[0.8pt]

\multicolumn{6}{l}{\textbf{Baseline}} \\
\addlinespace[1pt]
Rule-based
& 0B
& Qwen3-8B
& No
& 1$\times$
& 1$\times$ \\

\midrule[0.6pt]
\multicolumn{6}{l}{\textbf{Scalar Reward Models}} \\
\addlinespace[1pt]
Skywork-27B
& 27B
& Qwen3-8B
& No
& 1.125$\times$
&2.2$\times$ \\

Skywork-27B w/ responses
& 27B
& Qwen3-8B
& No
& 1.125$\times$
& 2.6$\times$ \\

Nemotron-70B
& 70B
& Qwen3-8B
& No
& 1.5$\times$
& 2.4$\times$ \\

Nemotron-70B w/ responses
& 70B
& Qwen3-8B
& No
& 1.5$\times$
& 3.5$\times$ \\

\midrule[0.6pt]
\multicolumn{6}{l}{\textbf{Static Rubric}} \\
\addlinespace[1pt]
RaR~\citep{gunjal2025rubrics}
& 235B
& Qwen3-8B
& Yes
& 1$\times^*$
& 2.5$\times^*$ \\

\midrule[0.6pt]
\multicolumn{6}{l}{\textbf{Dynamic Rubric}} \\
\addlinespace[1pt]
\rowcolor{mylightblue}
\textbf{DR-Generator-8B}
& 8B
& Qwen3-8B
& Yes
& 1$\times$
& 3.0$\times$ \\

\bottomrule[1.2pt]
\end{tabular}
}
\begin{minipage}{0.95\textwidth}
\footnotesize
\textit{$*$}
For RaR, the cost of using the 235B generator to produce rubrics for 36K prompts is \textbf{NOT} included.
\end{minipage}
\caption{
Computational overhead for policy optimization under different supervision methods.
Minimal GPU Allocation reports the minimum GPU resources that must be reserved for the training pipeline, and Actual GPU Hours reports the total end-to-end training cost in GPU hours.
For Actual GPU Hours, all pipelines are run with an additional $1\times$ auxiliary GPU allocation for acceleration.
For DynamicRubric, DR-Generator, DR-Verifier, and DR-Policy use the same backbone family, so the weights can be flushed and reloaded with SGLang without allocating additional GPU resources. The verifier in Static Rubric method can also be deployed in this way to reduce GPU allocation.
}

\label{tab:policy-compute-overhead}
\end{table*}

The GRPO objective is defined as
\begin{equation}
\small
\begin{aligned}
\mathcal{J}_{\mathrm{GRPO}}&(\theta) 
= \mathbb{E}_{q, \{o_i\}} \frac{1}{G} \sum_{i=1}^G \frac{1}{|o_i|} \sum_{t=1}^{|o_i|} \Bigg\{ \\
& \min \big[ r_{i,t}(\theta) \hat{A}_{i,t},\;
    \mathrm{clip}(r_{i,t}(\theta), 1-\varepsilon, 1+\varepsilon) \hat{A}_{i,t} \big]\\
& - \beta\,\mathbb{D}_{\mathrm{KL}}\big[\pi_\theta \,\|\, \pi_{\mathrm{ref}}\big] \Bigg\},
\end{aligned}
\end{equation}

where $r_{i,t}(\theta) = \frac{\pi_\theta(o_{i,t} \mid q, o_{i,<t})}{\pi_{\theta_{\mathrm{old}}}(o_{i,t} \mid q, o_{i,<t})}$ is the token-level probability ratio and $\beta$ controls the KL penalty strength with respect to a reference policy $\pi_{\mathrm{ref}}$.

In Section~\ref{sec:exp-setup}, we combine two objectives at the advantage level. Formally, that is
\begin{equation}
    A_{i,t} = \hat{A}^{disc}_{i,t} + \hat{A}^{anchor}_{i,t},
\end{equation}
where $\hat{A}^*_{i,t}$ is computed as
\begin{align}
    \hat{A}^{*}_{i,t}&= \frac{R^{*}_i - \mathrm{mean}(\{R^{*}_j\}_{j=1}^G)}{\mathrm{std}(\{R^{*}_j\}_{j=1}^G)},*\in\{anchor,disc\}
\end{align}

During training, we use the structured output mode of SGLang to constrain DR-Generator and DR-Verifier to produce the desired JSON format, simplifying extraction. Stage transitions are performed within an epoch.

\section{Computational Overhead}
\label{app:computational-overhead}
We report the computational overhead of different supervision methods during policy optimization in Table~\ref{tab:policy-compute-overhead}.
All numbers are normalized by the No-RM baseline using the same Qwen3-8B policy backbone.
Minimal GPU Allocation denotes the minimum amount of GPU resources that must be reserved to run the corresponding training pipeline.
Actual GPU Hours denotes the total end-to-end GPU-hour cost measured under a common accelerated setting, where every pipeline receives an additional $1\times$ auxiliary GPU allocation for inference or scoring acceleration.

Scalar reward models do not require a separate verifier, but larger reward models increase the minimum allocated GPU resources.
Conditioning scalar reward models on the full response set further increases total GPU hours because the scoring input becomes longer.
Static rubric supervision requires a verifier to apply generated rubrics, while its reported cost \textbf{excludes} the additional offline cost of generating rubrics for all training prompts with a 235B model.

DynamicRubric also requires rubric verification, but DR-Generator, DR-Verifier, and DR-Policy are instantiated from the same backbone family.
This allows the corresponding weights to be flushed and reloaded with SGLang rather than kept as separate persistent models.
As a result, DynamicRubric matches the No-RM-Rule-Based baseline in minimal GPU allocation while incurring additional total GPU hours from online rubric generation and verification.

\begin{table*}[t]
\centering
\small
\setlength{\tabcolsep}{4.8pt}
\renewcommand{\arraystretch}{1.12}
\resizebox{\textwidth}{!}{
\begin{tabular}{llcccccc}
\toprule[1.2pt]
\multirow{2}{*}{\textbf{Rubric Generator}}
& \multirow{2}{*}{\textbf{Size}}
& \textbf{JudgeBench}
& \textbf{Personalized-RB}
& \textbf{RM-Bench}
& \textbf{UltraFeedback}
& \textbf{AEOLLM}
& \textbf{LMSYS-Chat-1M} \\
&
& Acc.$\uparrow$
& Acc.$\uparrow$
& Acc.$\uparrow$
& Acc.$\uparrow$
& Acc.$\uparrow$
& Acc.$\uparrow$ \\
\midrule[0.8pt]

\multicolumn{8}{l}{\textbf{Verifier: Llama-3.1-8B-Instruct}} \\
\addlinespace[1pt]
Qwen3-8B zero-shot
& 8B
& 27.7
& 38.7
& 40.3
& 44.1
& 21.7
& 11.9 \\

Qwen3-32B zero-shot
& 32B
& 38.4
& 60.9
& 58.6
& 60.0
& 34.8
& 16.5 \\

\rowcolor{mylightblue}
\textbf{DR-Generator-8B}
& 8B
& \textbf{50.3}
& \textbf{68.6}
& \textbf{63.1}
& \textbf{63.0}
& \textbf{43.7}
& \textbf{26.1} \\

\midrule[0.6pt]
\multicolumn{8}{l}{\textbf{Verifier: Qwen3-8B}} \\
\addlinespace[1pt]
Qwen3-8B zero-shot
& 8B
& 33.1
& 52.5
& 49.9
& 51.8
& 18.2
& 11.0 \\

Qwen3-32B zero-shot
& 32B
& 42.8
& 67.8
& 62.8
& 62.6
& 25.3
& 14.1 \\

\rowcolor{mylightblue}
\textbf{DR-Generator-8B}
& 8B
& \textbf{54.9}
& \textbf{73.0}
& \textbf{67.0}
& \textbf{66.8}
& \textbf{46.5}
& \textbf{27.4} \\

\midrule[0.6pt]
\multicolumn{8}{l}{\textbf{Verifier: Qwen3-30B-A3B}} \\
\addlinespace[1pt]
Qwen3-8B zero-shot
& 8B
& 36.4
& 53.7
& 52.2
& 54.7
& 17.8
& 12.0 \\

Qwen3-32B zero-shot
& 32B
& 46.4
& 67.6
& 64.7
& 64.6
& 29.7
& 15.1 \\

\rowcolor{mylightblue}
\textbf{DR-Generator-8B}
& 8B
& \textbf{57.4}
& \textbf{73.3}
& \textbf{69.8}
& \textbf{68.7}
& \textbf{51.2}
& \textbf{28.0} \\

\midrule[0.6pt]
\multicolumn{8}{l}{\textbf{Verifier: Qwen3-32B}} \\
\addlinespace[1pt]
Qwen3-8B zero-shot
& 8B
& 35.8
& 54.6
& 50.7
& 53.3
& 21.3
& 12.7 \\

Qwen3-32B zero-shot
& 32B
& 46.7
& 69.5
& 65.4
& 65.5
& 32.2
& 16.5 \\

\rowcolor{mylightblue}
\textbf{DR-Generator-8B}
& 8B
& \textbf{57.2}
& \textbf{72.7}
& \textbf{68.0}
& \textbf{68.2}
& \textbf{52.5}
& \textbf{29.1} \\

\midrule[0.6pt]
\multicolumn{8}{l}{\textbf{Verifier: Qwen3-235B-A22B}} \\
\addlinespace[1pt]
Qwen3-8B zero-shot
& 8B
& 40.4
& 55.0
& 56.8
& 55.5
& 23.1
& 13.8 \\

Qwen3-32B zero-shot
& 32B
& 50.2
& 69.2
& 70.5
& 67.1
& 34.7
& 18.1 \\

\rowcolor{mylightblue}
\textbf{DR-Generator-8B}
& 8B
& \textbf{61.0}
& \textbf{75.5}
& \textbf{71.6}
& \textbf{69.9}
& \textbf{50.7}
& \textbf{30.1} \\

\bottomrule[1.2pt]
\end{tabular}
}
\caption{
Evaluator performance when dynamic rubric generators are paired with different verifiers.
Each block fixes the verifier used to apply generated rubrics and compares Qwen3-8B zero-shot, Qwen3-32B zero-shot, and the trained DR-Generator-8B.
Acc. denotes strict top-1 accuracy with ties excluded.
Within each verifier block, the best result is in \textbf{bold}.
}

\label{tab:verifier-robustness}
\end{table*}

\begin{table*}[t]
\centering
\small
\setlength{\tabcolsep}{6.0pt}
\renewcommand{\arraystretch}{1.12}
\resizebox{\textwidth}{!}{
\begin{tabular}{lclcccc}
\toprule[1.2pt]
\textbf{Supervision Method}
& \textbf{Supervisor Size}
& \textbf{Backbone}
& \textbf{AlpacaEval2}
& \textbf{ArenaHardv2.0}
& \textbf{WildBench}
& \textbf{WritingBench} \\
\midrule[0.8pt]

\multicolumn{7}{l}{\textbf{Baselines}} \\
\addlinespace[1pt]
Native
& -
& Qwen3-8B
& 22.7
& 3.6
& 38.4
& 64.4 \\

Native
& -
& Qwen3-32B
& 37.3
& 8.9
& 48.6
& 68.9 \\

\midrule[0.6pt]
\multicolumn{7}{l}{\textbf{Scalar Reward Models}} \\
\addlinespace[1pt]
Skywork-27B
& 27B
& Qwen3-8B
& 40.2
& 6.8
& 45.2
& 67.6 \\

Skywork-27B w/ responses
& 27B
& Qwen3-8B
& 36.5
& 5.4
& 43.5
& 65.4 \\

Nemotron-70B
& 70B
& Qwen3-8B
& 23.0
& 3.2
& 38.1
& 62.3 \\

Nemotron-70B w/ responses
& 70B
& Qwen3-8B
& 24.6
& 3.8
& 39.6
& 65.0 \\

\midrule[0.6pt]
\multicolumn{7}{l}{\textbf{Static Rubric}} \\
\addlinespace[1pt]
RaR~\citep{gunjal2025rubrics}
& 235B
& Qwen3-8B
& 54.1
& 10.5
& 47.6
& 69.2 \\

\midrule[0.6pt]
\multicolumn{7}{l}{\textbf{Dynamic Rubric}} \\
\addlinespace[1pt]
\rowcolor{mylightblue}
\textbf{DR-Generator-8B}
& 8B
& Qwen3-8B
& \textbf{67.1}
& \textbf{19.7}
& \textbf{51.6}
& \textbf{72.6} \\

\bottomrule[1.2pt]
\end{tabular}
}
\caption{
Policy performance on open-domain generation benchmarks across different supervision methods, evaluated with GPT-4.1 as the judge.
AlpacaEval2 and ArenaHardv2.0 report win rates, while WildBench and WritingBench report benchmark scores in their standard evaluation scales.
All scores are reported such that higher is better.
In each column, the best is in \textbf{bold}.
}
\label{tab:policy-results-gpt41}
\end{table*}

\section{Robustness Experiments and Analysis}
\subsection{Robustness to Verifier Choice}
\label{app:verifier-robustness}

DynamicRubric separates rubric generation from rubric verification: DR-Generator proposes weighted binary criteria, while DR-Verifier applies these criteria to candidate responses.
The main experiments use Qwen3-30B-A3B as the verifier.
To examine whether the gains depend on this specific verifier, we pair the same dynamic rubric generators with a range of frozen verifiers, including Llama-3.1-8B-Instruct and Qwen3 models of different scales.
For each verifier, we compare Qwen3-8B zero-shot rubric generation, Qwen3-32B zero-shot rubric generation, and the trained DR-Generator-8B.

Table~\ref{tab:verifier-robustness} shows that DR-Generator-8B remains the strongest rubric generator across all verifier choices.
The absolute scores vary with verifier strength, as expected, but the relative pattern is stable: trained dynamic rubric generation consistently improves over both zero-shot generators under every verifier.
This holds even when the verifier is a different model family, Llama-3.1-8B-Instruct, indicating that the learned rubric generator does not rely on a particular verifier or same-family verifier.
These results support the modularity of DynamicRubric: training improves the quality of generated rubrics in a way that transfers across the frozen models used to verify them.

\subsection{Robustness to LLM-as-Judge Choice}
\label{app:judge-robustness}

The main policy evaluation uses DeepSeek-V4-Flash as the automatic LLM-as-Judge.
To test whether the observed policy gains depend on this judge, we re-evaluate the same policy checkpoints using GPT-4.1 while keeping the benchmark prompts, generated responses, and evaluation protocols unchanged.
Table~\ref{tab:policy-results-gpt41} shows that DynamicRubric remains the strongest supervision method under GPT-4.1 evaluation.
It outperforms the native Qwen3-8B policy, scalar reward supervision, response-set-conditioned scalar variants, and static rubric supervision on all four benchmarks.
Compared with the 235B static rubric supervisor, DynamicRubric improves AlpacaEval2, ArenaHardv2.0, WildBench, and WritingBench by 13.0, 9.2, 4.0, and 3.4 points, respectively.
These results indicate that the downstream gains are robust to the choice of automatic judge.

\begin{table*}[t]
\centering
\small
\setlength{\tabcolsep}{6.0pt}
\renewcommand{\arraystretch}{1.12}
\begin{tabular}{llcccc}
\toprule[1.2pt]
\multirow{2}{*}{\textbf{Benchmark}}
& \multirow{2}{*}{\textbf{Metric}}
& \multicolumn{2}{c}{\textbf{Llama-3.1-8B-Instruct}}
& \multicolumn{2}{c}{\textbf{Qwen3-8B}} \\
\cmidrule(lr){3-4}
\cmidrule(lr){5-6}
&
& \textbf{Native}
& \textbf{DR-Generator}
& \textbf{Native}
& \textbf{DR-Generator} \\
\midrule[0.8pt]

JudgeBench
& Acc.$\uparrow$
& 34.8
& \textbf{46.8}
& 36.4
& \textbf{57.4} \\

Personalized-RB
& Acc.$\uparrow$
& 54.4
& \textbf{71.1}
& 53.7
& \textbf{73.3} \\

RM-Bench
& Acc.$\uparrow$
& 45.9
& \textbf{50.2}
& 52.2
& \textbf{69.8} \\

UltraFeedback
& Acc.$\uparrow$
& 56.2
& \textbf{64.2}
& 54.7
& \textbf{68.7} \\

AEOLLM
& Acc.$\uparrow$
& 19.2
& \textbf{44.8}
& 17.8
& \textbf{51.2} \\

LMSYS-Chat-1M
& Acc.$\uparrow$
& 13.3
& \textbf{24.7}
& 12.0
& \textbf{28.0} \\

\bottomrule[1.2pt]
\end{tabular}
\caption{
Evaluator performance of native and DynamicRubric-trained generators across Llama and Qwen backbones.
Native denotes the corresponding backbone used as a zero-shot dynamic rubric generator.
Acc. denotes strict top-1 accuracy, with tied top scores counted as incorrect.
For each backbone and metric, the better result is in \textbf{bold}.
}
\label{tab:backbone-evaluator-results}
\end{table*}
\begin{table*}[t]
\centering
\small
\setlength{\tabcolsep}{6.0pt}
\renewcommand{\arraystretch}{1.12}
\begin{tabular}{lcccc}
\toprule[1.2pt]
\multirow{2}{*}{\textbf{Benchmark}}
& \multicolumn{2}{c}{\textbf{Llama-3.1-8B-Instruct}}
& \multicolumn{2}{c}{\textbf{Qwen3-8B}} \\
\cmidrule(lr){2-3}
\cmidrule(lr){4-5}
&
\textbf{Native}
& \textbf{DR-Policy}
& \textbf{Native}
& \textbf{DR-Policy} \\
\midrule[0.8pt]

AlpacaEval2
& 15.3
& \textbf{28.7}
& 38.5
& \textbf{67.1} \\

ArenaHardv2.0
& 0.4
& \textbf{1.4}
& 9.0
& \textbf{21.0} \\

WildBench
& -17.1
& \textbf{-7.7}
& 16.4
& \textbf{30.7} \\

WritingBench
& 39.1
& \textbf{47.5}
& 57.8
& \textbf{64.3} \\

\bottomrule[1.2pt]
\end{tabular}
\caption{
Policy performance of native and DynamicRubric-optimized policies across Llama and Qwen backbones.
AlpacaEval2 and ArenaHardv2.0 report win rates, while WildBench and WritingBench report benchmark scores in their standard evaluation scales.
All scores are reported such that higher is better.
For each backbone and benchmark, the better result is in \textbf{bold}.
}
\label{tab:backbone-policy-results}
\end{table*}

\subsection{Cross-Backbone Robustness of DynamicRubric}
\label{app:additional-experiments-llama}
To evaluate robustness to the choice of model backbone, we repeat the full DynamicRubric pipeline with Llama-3.1-8B-Instruct~\citep{grattafiori2024llama3herdmodels}.
Both DR-Generator and DR-Policy are initialized from the native Llama-3.1-8B-Instruct checkpoint and trained with the same procedure used in the main experiments.
We compare the resulting DynamicRubric-trained models against their native counterparts under the same evaluation protocols.

Table~\ref{tab:backbone-evaluator-results} shows that DynamicRubric training consistently improves the Llama-based evaluator.
DR-Generator outperforms the native generator on every pairwise and listwise benchmark.
This confirms that the evaluator-side benefit of response-set-conditioned rubric training is not specific to the Qwen backbone.

Table~\ref{tab:backbone-policy-results} shows that the Llama-based DR-Policy also improves over the native policy on all four open-domain generation benchmarks.
The gains cover both win-rate benchmarks and benchmark-score evaluations, indicating that DynamicRubric supervision remains effective after changing the policy backbone.
Together with the evaluator results, these experiments demonstrate cross-backbone robustness of the full bilevel DynamicRubric pipeline.

\begin{table*}[t]
\centering
\begin{tabular}{lcccc}
\toprule
Generator & Response A & Response B & 5-seed nDCG & 5-seed Acc. \\
\midrule
$G_0$ & 15 & 15 & 0.8155 & 0.0 \\
$G_2$ & 53 & 24 & 1.0000 & 1.0 \\
\bottomrule
\end{tabular}
\caption{
Case-study evaluator scores and aggregate ranking metrics.
Response A is the annotated preferred response.
Seed-0 scores are raw weighted sums from one generated rubric set.
Aggregate metrics are computed over five independently generated rubric sets with shuffled rubric orders.
Tied top scores are counted as incorrect for Acc.
}
\label{tab:case-study}
\end{table*}
\section{Rubric Quality Meta-Evaluation}
\label{app:rubric-quality}

We further conduct a rubric-level meta-evaluation to examine whether DynamicRubric training improves the quality of generated rubric lists beyond their downstream ranking accuracy.
We randomly sample 100 instances from Personalized-RewardBench and 100 instances from UltraFeedback.
For each sampled instance, GPT-4.1 is given the user prompt, the response pool, and two anonymized weighted rubric lists generated before and after DynamicRubric training, respectively, by the native Qwen3-8B generator and the trained DR-Generator-8B.
The judge compares the two rubric lists along task alignment, local discrimination, response awareness, specificity, weighting quality, and overall quality.

\begin{figure}[t]
    \centering
    \includegraphics[width=\linewidth]{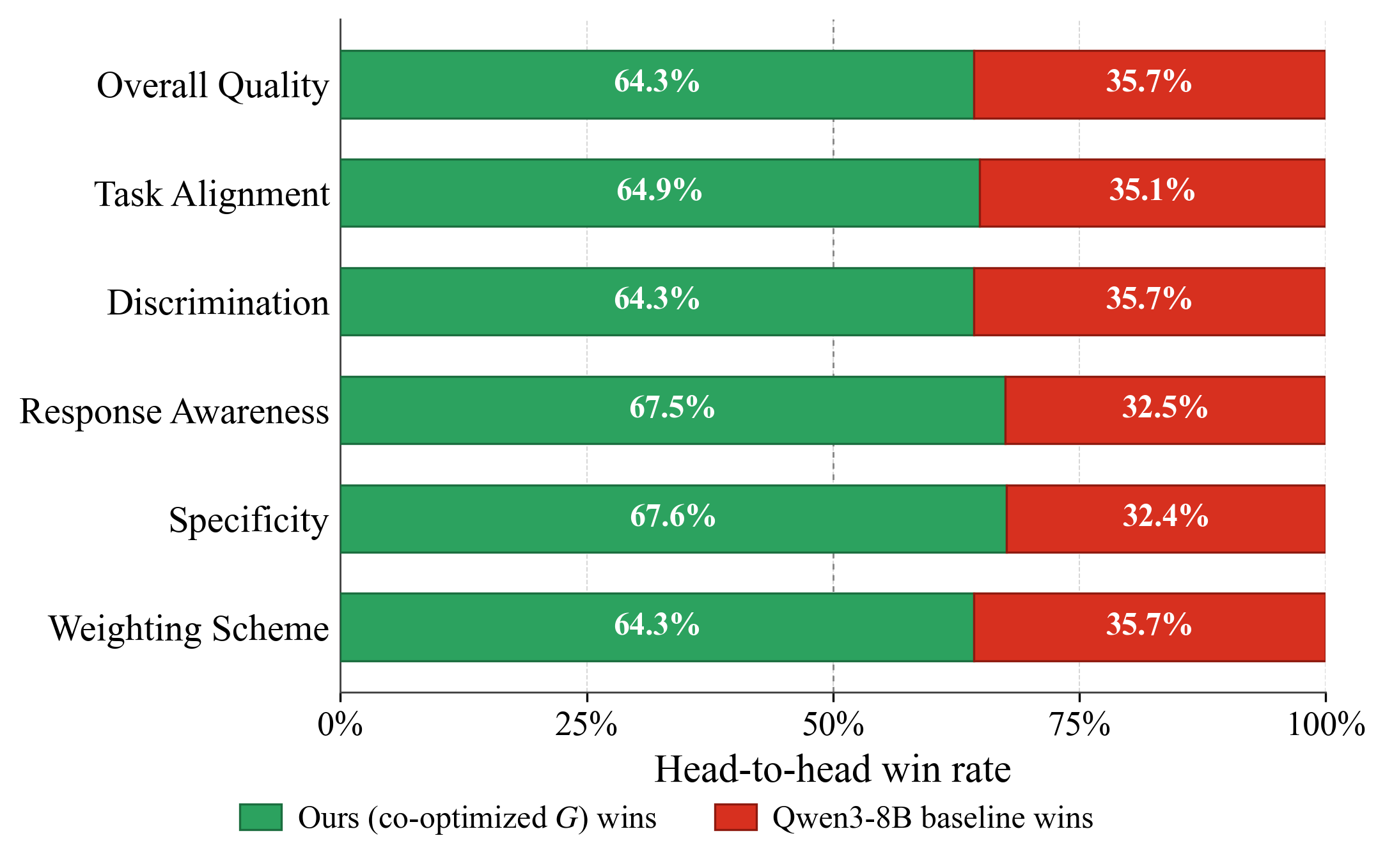}
    \caption{
    GPT-4.1 meta-evaluation of rubric quality.
    We report the pairwise win rate of DR-Generator-8B against the native Qwen3-8B generator on each rubric-quality dimension.
    }
    \label{fig:rubric-quality}
\end{figure}

Figure~\ref{fig:rubric-quality} reports the fraction of pairwise comparisons in which GPT-4.1 prefers the DynamicRubric-trained generator.
DR-Generator-8B is preferred over the native generator across all dimensions, with a $64.3\%$ overall win rate.
The strongest gains appear in response awareness ($67.5\%$) and specificity ($67.6\%$), indicating that DynamicRubric training makes rubric generation more sensitive to the decisive differences within the observed response set while maintaining concrete and assessable criteria.
The judge also prefers DR-Generator-8B on task alignment ($64.9\%$), local discrimination ($64.3\%$), and weighting quality ($64.3\%$).
These results provide complementary qualitative evidence that DynamicRubric improves the rubric generator in the intended direction: generating criteria that are more discriminative, response-aware, and useful for evaluator-guided policy optimization.

\section{Case Study}
\label{app:case-study}

We present a representative JudgeBench-GPT example to illustrate how DynamicRubric training changes response-set-conditioned rubric generation.
Both generators receive only the user prompt and the unordered candidate response set.
The annotated preference is withheld during rubric generation and used only for evaluation.
The zero-shot Qwen3-8B generator $G_0$ and the DynamicRubric-trained generator $G_2$ are evaluated with the same frozen DR-Verifier.
For each generated rubric set, DR-Verifier executes every rubric item as a binary yes/no check, and the weighted outcomes are aggregated into response-level evaluator scores.
The score table reports one illustrative seed-0 run, while aggregate metrics are computed over five independently generated rubric sets with shuffled rubric orders.
\begin{promptbox}{Case-study input observed by both generators}
Benchmark: JudgeBench-GPT

User prompt:
Estimate the angular momentum of Hurricane Florence at landfall, choose one option from A-J, and repeat the selected letter five times.

Candidate response set, unordered:

Response A:
Uses L = I omega with a disk-like moment of inertia.
Specifies numerical estimates for mass, radius, and angular velocity.
Obtains L about 3e19 kg m^2 / s and selects option B.

Response B:
Uses L = m r^2 omega.
Uses an angular velocity of about 1e-1 rad/s.
Obtains L about 1e23 kg m^2 / s and selects option A.
\end{promptbox}

\begin{promptbox}{Rubrics generated by zero-shot Qwen3-8B generator $G_0$}
1. The response provides the answer letter and repeats it five times.
2. The response includes step-by-step reasoning.
3. The response uses the correct angular-momentum formula.
4. The response stays focused on the question.
\end{promptbox}

\begin{promptbox}{Representative rubrics generated by DynamicRubric-trained generator $G_2$}
1. The response mentions the exact values used for mass, radius, or angular velocity.
2. The response compares the estimated angular momentum with the provided answer options.
3. The response mentions central pressure or wind speed as part of the estimation.
4. The response states the disk-like approximation for the hurricane.
5. The response discusses the proportionality constant in the moment of inertia.
6. The response computes angular velocity from tangential velocity and radius.
7. The response estimates mass and radius using typical hurricane-scale values.
8. The response computes both moment of inertia and angular momentum.
\end{promptbox}

The zero-shot generator $G_0$ produces valid but weakly discriminative criteria.
Its rubrics check answer formatting, step-by-step presentation, formula usage, and topical relevance.
Both candidate responses satisfy these broad criteria: each gives an option in the required format, presents a calculation, invokes an angular-momentum formula, and stays on topic.
The frozen DR-Verifier therefore assigns identical weighted scores to the two responses, which removes the unique top-ranked response and gives zero strict top-1 accuracy across all five seeds.

The DynamicRubric-trained generator $G_2$ produces criteria that better reflect the quality differences within the observed response set.
Beyond checking formula presence, its rubrics examine whether the response specifies physically relevant quantities, estimates angular velocity from wind speed and radius, states an explicit moment-of-inertia model, and compares the resulting order of magnitude with the answer options.
These criteria favor Response A, which carries out a coherent order-of-magnitude estimate, over Response B, whose final choice is driven by an implausibly large angular-velocity assumption.
With the same frozen DR-Verifier, the generated rubric set induces a clear response-level score gap and ranks the annotated preferred response first in every seed.

This case illustrates the effect of learned dynamic rubrics: after DynamicRubric training, DR-Generator uses the observed alternatives to generate criteria that are both task-valid and locally discriminative for the current comparison.
\section{Potential Risks}
DynamicRubric optimizes policies with rubric-based automatic supervision, so policies may learn to satisfy verifier-specific preferences or rubric-compliance patterns that do not fully match broader human judgments.
Generated rubrics may also inherit biases or blind spots from the generator and verifier, and repeated optimization can amplify such patterns.
Although our robustness checks and qualitative inspection provide partial safeguards, they do not replace large-scale human validation.
Future work can study verifier ensembles, human calibration, and auditing procedures for bias and reward-hacking behavior before deployment.

\clearpage

\onecolumn

\section{A Residual Analysis of Discriminability and Anchor Rewards}
\label{app:alignment-guarantee}
\textbf{Following the formatting allowance for math-heavy appendices, this section is presented in single-column format because it consists primarily of formal proofs and derivations.}

This appendix derives the expected relative score gap bound used in Section~\ref{sec:dynamic-rubrics}.
The analysis separates the roles of the two Generator rewards.
The discriminability reward encourages generated rubric items to separate responses in current-policy-induced response sets.
The anchor reward calibrates the direction of these separations using ranked anchor data.
The analysis does not require every generated rubric item to be aligned with the reference order or covered by the ordered-pair distribution; misalignment and insufficient coverage appear as explicit residual terms.

We first derive a residual lower bound that holds for arbitrary generated rubrics.
Under an aggregate anchor-calibration condition, this residual bound yields constants
$c_{\mathrm{disc}},c_{\mathrm{anchor}}>0$ and a round-dependent residual $\epsilon_t$ independent of $\phi$ such that
\begin{equation}
\label{eq:app-main-guarantee}
\Delta_{\mathrm{gap}}(E_\phi;\theta_{t-1})
\ge
c_{\mathrm{disc}}\,
\mathcal J_{\mathrm{disc}}^{t}(\phi;\theta_{t-1})
+
c_{\mathrm{anchor}}\,
\mathcal J_{\mathrm{anchor}}^{t}(\phi;\theta_{t-1})
+
\epsilon_t.
\end{equation}
Since $\epsilon_t$ is independent of $\phi$ within the evaluator-update candidate class, it does not affect the evaluator update.

Fix a prompt $x$ and a conditioning response set
\begin{equation}
\label{eq:app-fixed-c}
\mathcal C=\{y_1,\dots,y_K\}.
\end{equation}
Let
\begin{equation}
\label{eq:app-rubric-list}
\mathcal R_\phi(x,\mathcal C)=\{(r_m,w_m)\}_{m=1}^{M}
\end{equation}
be the rubrics generated from $\mathcal C$.
For notational simplicity, we suppress the dependence of rubric items, weights, and verifier outcomes on $(x,\mathcal C,\phi)$ whenever $x$ and $\mathcal C$ are fixed.

Let the normalized rubric weights be
\begin{equation}
\label{eq:app-normalized-weights}
a_m
=
\frac{w_m}{\sum_{\ell=1}^{M}w_\ell},
\qquad
\sum_{m=1}^{M}a_m=1.
\end{equation}
For rubric item $r_m$, denote the DR-Verifier outcome on response $y$ by
\begin{equation}
\label{eq:app-verifier-outcome}
v_m(y)
=
V(x,y,r_m)
\in\{0,1\}.
\end{equation}
The response-level evaluator induced by the rubric set is
\begin{equation}
\label{eq:app-evaluator}
E_\phi(x,y\mid\mathcal C)
=
\sum_{m=1}^{M}a_m v_m(y).
\end{equation}
The discriminability reward can be written as
\begin{equation}
\label{eq:app-disc-reward}
R_{\mathrm{disc}}(x,\mathcal C;\phi)
=
\sum_{m=1}^{M}
a_m\bar v_m(1-\bar v_m),
\qquad
\bar v_m
=
\frac1K
\sum_{i=1}^{K}
v_m(y_i).
\end{equation}

Let $P(\cdot\mid x,\mathcal C)$ be the ordered-pair distribution from Section~\ref{sec:relative-gap}, instantiated on the conditioning response set.
For each rubric item $m$, define its signed gap contribution under $P$ as
\begin{equation}
\label{eq:app-a-m}
A_m
=
\mathbb E_{(y^+,y^-)\sim P(\cdot\mid x,\mathcal C)}
\left[
v_m(y^+)-v_m(y^-)
\right],
\end{equation}
and define its split mass as
\begin{equation}
\label{eq:app-b-m}
B_m
=
\mathbb E_{(y^+,y^-)\sim P(\cdot\mid x,\mathcal C)}
\left[
|v_m(y^+)-v_m(y^-)|
\right].
\end{equation}
Here, $B_m$ measures how often rubric item $m$ separates ordered pairs induced by $\mathcal C$, while $A_m$ is the signed version that measures whether these separations follow the reference ordering.

\paragraph{Directional residual.}
For any $\gamma>0$, define the directional residual of rubric item $m$ as
\begin{equation}
\label{eq:directional-residual}
\xi_m^\gamma
=
\left[\gamma B_m-A_m\right]_+,
\end{equation}
where $[z]_+=\max\{z,0\},\,z\in \mathbb R$.
Then, by construction,
\begin{equation}
\label{eq:directional-residual-bound}
A_m
\ge
\gamma B_m-\xi_m^\gamma.
\end{equation}
The residual $\xi_m^\gamma$ measures the extent to which the split induced by rubric item $m$ fails to follow the reference ordering on the evaluated ordered pairs.

\paragraph{Coverage residual.}
For any $\rho>0$, define the aggregate coverage residual as
\begin{equation}
\label{eq:coverage-residual}
\zeta_{\mathrm{cov}}^\rho(x,\mathcal C;\phi)
=
\left[
\rho R_{\mathrm{disc}}(x,\mathcal C;\phi)
-
\sum_{m=1}^{M}a_mB_m
\right]_+.
\end{equation}
Then,
\begin{equation}
\label{eq:coverage-residual-bound}
\sum_{m=1}^{M}a_mB_m
\ge
\rho R_{\mathrm{disc}}(x,\mathcal C;\phi)
-
\zeta_{\mathrm{cov}}^\rho(x,\mathcal C;\phi).
\end{equation}
This residual is small when the ordered-pair distribution places nontrivial mass on response pairs separated by the generated rubric items.

\paragraph{Lemma 1: Fixed-input residual lower bound.}
\label{lemma:fixed-input-bound}
For fixed $x$ and $\mathcal C$ and any $\gamma,\rho>0$,
\begin{equation}
\label{eq:fixed-input-bound}
\begin{aligned}
&
\mathbb E_{(y^+,y^-)\sim P(\cdot\mid x,\mathcal C)}
\left[
E_\phi(x,y^+\mid\mathcal C)
-
E_\phi(x,y^-\mid\mathcal C)
\right]
\\
&\ge
\gamma\rho R_{\mathrm{disc}}(x,\mathcal C;\phi)
-
\sum_{m=1}^{M}a_m\xi_m^\gamma
-
\gamma
\zeta_{\mathrm{cov}}^\rho(x,\mathcal C;\phi).
\end{aligned}
\end{equation}

\paragraph{Proof.}
Fix $x$, $\mathcal C$, and the generated rubric list.
Define the expected evaluator gap over ordered pairs as
\begin{equation}
\label{eq:app-g-e}
G_E
=
\mathbb E_{(y^+,y^-)\sim P(\cdot\mid x,\mathcal C)}
\left[
E_\phi(x,y^+\mid\mathcal C)
-
E_\phi(x,y^-\mid\mathcal C)
\right].
\end{equation}
By the linearity of the evaluator in Equation~\ref{eq:app-evaluator},
\begin{equation}
\label{eq:app-ge-linear}
\begin{aligned}
G_E
&=
\sum_{m=1}^{M}
a_m
\mathbb E_{(y^+,y^-)\sim P(\cdot\mid x,\mathcal C)}
\left[
v_m(y^+)-v_m(y^-)
\right]
\\
&=
\sum_{m=1}^{M}a_mA_m.
\end{aligned}
\end{equation}
Using Equation~\ref{eq:directional-residual-bound} and $a_m\ge 0$,
\begin{equation}
\label{eq:app-use-directional-residual}
G_E
\ge
\gamma
\sum_{m=1}^{M}a_mB_m
-
\sum_{m=1}^{M}a_m\xi_m^\gamma.
\end{equation}
Using Equation~\ref{eq:coverage-residual-bound},
\begin{equation}
\label{eq:app-use-coverage-residual}
G_E
\ge
\gamma\rho R_{\mathrm{disc}}(x,\mathcal C;\phi)
-
\gamma
\zeta_{\mathrm{cov}}^\rho(x,\mathcal C;\phi)
-
\sum_{m=1}^{M}a_m\xi_m^\gamma.
\end{equation}
Substituting the definition of $G_E$ in Equation~\ref{eq:app-g-e} proves Lemma~1.
\hfill$\square$

Define the expected residual during the evaluator update at round $t$ as
\begin{equation}
\label{eq:eta-res}
\eta_{\mathrm{res}}^{t,\gamma,\rho}(\phi)
=
\mathbb E_{x\sim\mathcal D}
\mathbb E_{\mathcal C\sim\pi_{\theta_{t-1}}(\cdot\mid x)}
\left[
\sum_{m=1}^{M}a_m(x,\mathcal C)\xi_m^\gamma(x,\mathcal C)
+
\gamma
\zeta_{\mathrm{cov}}^\rho(x,\mathcal C;\phi)
\right].
\end{equation}
This quantity collects the residual direction error and the residual coverage error on the response-set distribution used for the evaluator update.

\paragraph{Aggregate anchor calibration.}
\label{assump:anchor-control}
For the chosen $\gamma,\rho>0$, suppose there exist a constant $\beta>0$ and a round-dependent residual $\epsilon_t$ independent of $\phi$ within the evaluator-update candidate class at round $t$ such that
\begin{equation}
\label{eq:assump-anchor-control}
-\eta_{\mathrm{res}}^{t,\gamma,\rho}(\phi)
\ge
\beta\,
\mathcal J_{\mathrm{anchor}}^{t}(\phi;\theta_{t-1})
+
\epsilon_t.
\end{equation}
This condition states that the anchor ranking objective controls the aggregate residual left by discriminability-only training.
For the Bradley--Terry-style anchor reward in Equation~\ref{eq:anchor-reward}, $\mathcal J_{\mathrm{anchor}}^{t}$ is non-positive and increases toward zero as the induced score gaps better agree with the anchor ranking direction.
Thus, increasing the anchor objective raises the lower bound by reducing the aggregate residual.

\paragraph{Theorem 1: Expected relative score gap guarantee.}
\label{thm:expected-gap-guarantee}
For any $\gamma,\rho>0$, the expected relative score gap satisfies
\begin{equation}
\label{eq:residual-gap-bound}
\Delta_{\mathrm{gap}}(E_\phi;\theta_{t-1})
\ge
\gamma\rho
\mathcal J_{\mathrm{disc}}^{t}(\phi;\theta_{t-1})
-
\eta_{\mathrm{res}}^{t,\gamma,\rho}(\phi).
\end{equation}
If the aggregate anchor calibration condition in Equation~\ref{eq:assump-anchor-control} holds, then there exist constants
$c_{\mathrm{disc}},c_{\mathrm{anchor}}>0$ and a round-dependent residual $\epsilon_t$ independent of $\phi$ such that
\begin{equation}
\label{eq:theorem-guarantee}
\Delta_{\mathrm{gap}}(E_\phi;\theta_{t-1})
\ge
c_{\mathrm{disc}}\,
\mathcal J_{\mathrm{disc}}^{t}(\phi;\theta_{t-1})
+
c_{\mathrm{anchor}}\,
\mathcal J_{\mathrm{anchor}}^{t}(\phi;\theta_{t-1})
+
\epsilon_t.
\end{equation}
In particular, one can take
\begin{equation}
\label{eq:c-disc-c-anchor}
c_{\mathrm{disc}}=\gamma\rho,
\qquad
c_{\mathrm{anchor}}=\beta.
\end{equation}

\paragraph{Proof.}
Taking expectation in Lemma~1 over the evaluator-update sampling process,
$x\sim\mathcal D$ and
$\mathcal C\sim\pi_{\theta_{t-1}}(\cdot\mid x)$,
the left-hand side becomes
$\Delta_{\mathrm{gap}}(E_\phi;\theta_{t-1})$.
The first term on the right-hand side becomes
\begin{equation}
\label{eq:theorem-disc-term}
\gamma\rho
\mathbb E_{x\sim\mathcal D}
\mathbb E_{\mathcal C\sim\pi_{\theta_{t-1}}(\cdot\mid x)}
\left[
R_{\mathrm{disc}}(x,\mathcal C;\phi)
\right]
=
\gamma\rho
\mathcal J_{\mathrm{disc}}^t(\phi;\theta_{t-1}).
\end{equation}
The remaining terms become
$\eta_{\mathrm{res}}^{t,\gamma,\rho}(\phi)$ by Equation~\ref{eq:eta-res}.
Therefore,
\begin{equation}
\label{eq:gap-disc-minus-residual}
\Delta_{\mathrm{gap}}(E_\phi;\theta_{t-1})
\ge
\gamma\rho
\mathcal J_{\mathrm{disc}}^t(\phi;\theta_{t-1})
-
\eta_{\mathrm{res}}^{t,\gamma,\rho}(\phi),
\end{equation}
which proves Equation~\ref{eq:residual-gap-bound}.
Applying the aggregate anchor calibration condition in Equation~\ref{eq:assump-anchor-control} gives
\begin{equation}
\label{eq:theorem-final-substitution}
\begin{aligned}
\Delta_{\mathrm{gap}}(E_\phi;\theta_{t-1})
&\ge
\gamma\rho
\mathcal J_{\mathrm{disc}}^t(\phi;\theta_{t-1})
+
\beta
\mathcal J_{\mathrm{anchor}}^{t}(\phi;\theta_{t-1})
+
\epsilon_t.
\end{aligned}
\end{equation}
Setting $c_{\mathrm{disc}}=\gamma\rho$ and $c_{\mathrm{anchor}}=\beta$ proves Equation~\ref{eq:theorem-guarantee}.
\hfill$\square$
\clearpage
\section{Prompt Template}
We provide our prompt templates here.

\begin{promptbox}{Prompt template for Policy.}
Answer the user prompt.
Please reason step by step in <think>, write the complete user-facing answer in <answer>, and put the same final answer or conclusion from <answer> in minimal parseable form in <result>.

## User Prompt:
[user_prompt_replace]

## Output Format:
<think>
Your reasoning process.
</think>

<answer>
Your complete user-facing answer: include the final answer and the key supporting steps that justify it
</answer>

<result>
Your minimal final result or concise conclusion.
</result>
\end{promptbox}

\begin{promptbox}{Prompt template for Generator.}
You are an Evaluation Criteria Designer. Your task is to generate a list of binary ("yes"/"no") rubrics with weights to evaluate responses to a specific user prompt.

You will be provided with an unordered pool of actual model responses. Use this pool by comparing the responses to notice possible quality dimensions that matter for answering the user prompt. The response pool is only there to help you notice useful patterns and differences. If you notice a specific detail, example, entity, or mistake in the responses, first think about whether it reflects a broader quality dimension that is actually needed or meaningfully helpful for answering the user's request. If so, abstract it into a broader evaluative concept rather than copying it directly into a rubric. Do not treat the stronger responses in the pool as perfect; they may still contain errors or unnecessary details. Do not assume that every feature of stronger responses is equally necessary.

## Rubric Design Requirements:
1. Focus on the user's request:
   - Design rubrics to evaluate how well a response answers the user's actual request.
   - Use the response pool to notice which qualities help a response satisfy the user's request better.
   - Do not hardcode specific facts, numbers, names, entities, or other concrete details into your rubrics unless they are explicitly provided in the user prompt. Only use such details if you can abstract them into a broader evaluative concept that matters for answering the user's request.
   - Avoid rubrics that require external factual knowledge to verify.

2. Evaluate constraint adherence:
   - If the user prompt asks for specific formats, lengths, styles, or output structures, create rubrics to strictly check those constraints.

3. Atomicity and absolute binary (no "partially correct"):
   - Each rubric must evaluate exactly one distinct aspect.
   - Rubrics must be strictly binary (true or false).
   - If a response could reasonably satisfy only part of a rubric, that rubric is invalid and must be split or rewritten.
   - Bad example: "The response explains the cause and gives a solution."

4. Specificity and objectivity:
   - Prefer rubrics anchored to clear response features, explicit logical steps, or specific constraints requested by the user.
   - Make sure the boundary of the rubric is clear enough that different evaluators would usually agree on "yes" or "no".
   - Avoid vague descriptions.
   - Do not write rubrics around highly specific mistakes that appear in only one or two responses unless they reflect a broader quality dimension that would matter in other responses too.
   - Bad example: "The response is well-written and helpful."

5. Focus on substance:
   - Do not include rubrics that mainly distinguish responses by formatting or other presentation choices unless the user prompt explicitly requires them.
   - If the user did not ask for a specific format, a format-based rubric should usually be omitted.

6. Positive framing ("yes" = good):
   - Frame all rubrics so that a "yes" evaluation means the response is good, and "no" means it is bad.
   - Do not use double negatives.

7. No reference to specific responses:
   - Rubrics must not refer to any specific response.
   - Each rubric should be written as a general standard that can be applied to evaluate any response independently.

8. Mutually exclusive and self-contained:
   - Ensure each rubric evaluates a distinct aspect.
   - Each rubric should contain all the information needed to evaluate a response.
   - If one broad rubric and one narrower rubric are really checking the same thing, keep only one of them or rewrite them so they check clearly different things.
   - Do not treat two rubrics as overlapping just because they are related. Merge them only if a response would usually pass or fail them for the same reason.

9. Assigning weights (1-5 scale):
   - Assign an integer weight between 1 and 5 to each rubric based on how necessary that rubric is for a strong response to the user's request.
   - Essential (Weight 5): Core task requirements, critical user constraints, or safety-related requirements; if missing, the response is seriously flawed or invalid.
   - Important (Weight 3-4): Key reasoning, completeness, or clarity; these strongly affect response quality.
   - Optional (Weight 1-2): Nice-to-have improvements; these can distinguish an excellent response from a merely good one, but their absence should not make a response poor.
   - Do not assign a high weight merely because a rubric is common among stronger responses. A rubric can be useful for distinguishing stronger responses without being highly necessary.
   - Do not discard a rubric only because it is not essential. If it adds a different and useful quality dimension, keep it at a lower weight.
   - When the task supports it, include a mix of essential, important, and optional rubrics to help distinguish weaker, acceptable, and stronger responses.

10. The number of rubrics:
   - The appropriate number of rubrics depends on the task, but the rubric list should include enough independent rubrics to cover the meaningful quality dimensions needed for a discriminating evaluation.

## Rubric Design Workflow:
- First look at the user prompt on its own to identify the core task, any explicit constraints, and the minimum requirements for a response to count as a valid answer.
- Then consider the response pool together with the user prompt to identify the different quality dimensions that matter for this task.
- Use the response pool to notice useful quality differences, but do not let it override the user's actual request or constraints.
- When you notice a specific detail in the responses, abstract it into a broader quality dimension if that dimension is genuinely useful for evaluation.
- Decide for each quality dimension whether it is essential, important, or optional.
- After that, merge or remove rubrics only when they are really checking the same thing or when one adds almost no useful new information.
- Write each rubric as one complete declarative sentence.

## User Prompt:
[user_prompt_replace]

## Response List:
[response_list_replace]

## Output Format:
Please reason step by step, and then return only a valid JSON list inside <answer> and </answer> at the end of your output. The format of the <answer> block must be exactly:
<answer>
[
   {"rubric": "<rubric here>", "weight": <integer between 1 and 5>},
   {"rubric": "<rubric here>", "weight": <integer between 1 and 5>}
]
</answer>
\end{promptbox}

\begin{promptbox}{Prompt template for Verifier.}
You will be given:
1. A user prompt.
2. A list of rubrics, each with a unique ID.
3. A response to the user prompt.

Your task is to evaluate the response against each rubric independently and decide whether each rubric is satisfied.

## Scoring Rules:
- Answer "yes" if the response clearly meets the rubric.
- Answer "no" if the response does not meet the rubric. Partial matches should be scored as "no".
- If a rubric is poorly written and cannot be reasonably answered with "yes" or "no", it is a bad rubric - automatically judge it as "no".
- You must evaluate EVERY rubric provided in the list. Do not skip any.

## User Prompt:
[user_prompt_replace]

## Rubrics:
[rubrics_replace]

## Response:
[response_replace]

## Output Format:
Please return only a valid JSON list inside <answer> and </answer>. You MUST generate exactly one JSON object for EVERY rubric ID provided in the input. The number of items in your output list must exactly match the number of rubrics. Do not miss any ID, do not repeat any ID, and do not change the IDs. For each rubric, include all three fields: "id", "reason", and "judgement". Keep the same rubric ID order as the input list. Go through the IDs one by one. The "reason" should be brief and based only on what is clearly stated in the response. Do not guess missing facts, use outside knowledge, or revise your own answer inside the reason. If something is missing, simply say it is missing. Do not forget to write the "judgement" field for every item. The format must be exactly:
<answer>
[
   {"id": 1, "reason": "explain your judgement", "judgement": "yes"},
   {"id": 2, "reason": "explain your judgement", "judgement": "no"}
]
</answer>
Do not output anything else.
\end{promptbox}

\end{document}